\documentclass[10pt,twocolumn,letterpaper]{article}

\usepackage[pagenumbers]{cvpr} 

\usepackage[dvipsnames]{xcolor}

\usepackage{multirow}

\definecolor{cvprblue}{rgb}{0.21,0.49,0.74}
\usepackage[pagebackref,breaklinks,colorlinks,citecolor=cvprblue]{hyperref}

\def\paperID{193} 
\def\confName{3DV\xspace}
\def\confYear{2025\xspace}

\title{MLI-NeRF: Multi-Light Intrinsic-Aware Neural Radiance Fields}

\author{Yixiong Yang$^1$, Shilin Hu$^2$, Haoyu Wu$^2$, Ramon Baldrich$^1$, Dimitris Samaras$^2$, Maria Vanrell$^1$\\
$^1$ Universitat Autonoma de Barcelona, $^2$ Stony Brook University\\
{\tt\small \{yixiong, ramon, maria\}@cvc.uab.cat, \{shilhu, haoyuwu, samaras\}@cs.stonybrook.edu}
}

\begin{document}

\maketitle
\begin{abstract}

Current methods for extracting intrinsic image components, such as reflectance and shading, primarily rely on statistical priors. These methods focus mainly on simple synthetic scenes and isolated objects and struggle to perform well on challenging real-world data. To address this issue, we propose MLI-NeRF, which integrates \textbf{M}ultiple \textbf{L}ight information in \textbf{I}ntrinsic-aware \textbf{Ne}ural \textbf{R}adiance \textbf{F}ields. By leveraging scene information provided by different light source positions complementing the multi-view information, we generate pseudo-label images for reflectance and shading to guide intrinsic image decomposition without the need for ground truth data. Our method introduces straightforward supervision for intrinsic component separation and ensures robustness across diverse scene types. We validate our approach on both synthetic and real-world datasets, outperforming existing state-of-the-art methods. Additionally, we demonstrate its applicability to various image editing tasks. The code and data are publicly available at \url{https://github.com/liulisixin/MLI-NeRF}. 

\end{abstract}    
\vspace{-16pt}
\section{Introduction}
\vspace{-2pt}
Neural radiance fields (NeRF) have enabled significant strides in novel view synthesis (NVS) \cite{mildenhall2021nerf, li2023neuralangelo} including efforts towards scene editing \cite{wang2023udcnerf}, such as recoloring \cite{Ye2023IntrinsicNeRF}  and relighting \cite{ling2022shadowneus, zeng2023nrhints}. Scene editing becomes easier when the scene can be decomposed into editable sub-attributes.
There are two related approaches to scene editing \cite{garces2022survey}: inverse rendering and intrinsic image decomposition.

The first approach \cite{Jin2023TensoIR, zhang2022invrender, yang2023sireir, zhang2021nerfactor} integrates inverse rendering with neural rendering methods for scene decomposition. They often employ the BRDF \cite{burley2012physically}
to model material properties and jointly optimize geometry, materials, and environmental lighting. However, inverse rendering presents a highly ill-posed challenge: separating material properties and illumination in images often yields ambiguous results, and tracing light within scenes is computationally intensive. These factors limit inverse rendering to object-specific scenarios. The second approach \cite{Ye2023IntrinsicNeRF, careagaIntrinsic, dasPIENet}, based on intrinsic image decomposition \cite{barrow1978recovering}, aims to provide an explainable representation of a scene in terms of components such as reflectance and shading. In general, intrinsic image decomposition is more applicable to a broader range of scenarios, including individual objects and more complex scenes with backgrounds. 
While IntrinsicNeRF \cite{Ye2023IntrinsicNeRF} has pioneered the integration of intrinsic decomposition within NeRF, it has not fully leveraged the 3D information available through neural rendering. 

Mineralogists illuminate their specimens from different angles to reveal their features. Similarly, varying the light source position is essential for uncovering the intrinsic details of a scene. 
We aim to enhance intrinsic decomposition quality and expand scene editing capabilities by leveraging multiple light sources to build an intrinsic-aware NeRF.
The connection between varying lighting conditions and intrinsic decomposition has been discussed for 2D images \cite{BigTimeLi18,lettry2018unsupervised}, but not yet in neural rendering, even though there is interest in relighting using neural rendering \cite{zeng2023nrhints, rudnev2022nerfosr}. 

In this paper, we introduce \textbf{MLI-NeRF}, a two-stage method to learn an intrinsic-aware NeRF.
In Stage 1, we extend NeRF to incorporate light position information and learn a relightable scene using images captured from various camera angles and light source positions. 
In the subsequent post-processing, we begin by obtaining normals and light visibility maps for images under multiple lighting conditions using the model from Stage 1 and sphere tracing \cite{chen2022tracing}. 
We then generate pseudo shadings from the normals, light rays, and light visibility maps. 
Finally, for each camera pose, pseudo reflectance is generated by combining cues from multiple lighting conditions.
In Stage 2, we make our model intrinsic-aware by introducing additional modules for reflectance and shading while restricting light position input to the shading module only, ensuring the independence of the reflectance and light. 
In this paper, we forego potentially oversimplifying statistical constraints on various illumination-related factors in recent work \cite{Ye2023IntrinsicNeRF, lettry2018unsupervised, careagaIntrinsic, li2018learning} to instead use the physics-based disentanglement of reflectance and shading and achieve high-quality results.

As illustrated in \cref{teaser}, our method achieves high-quality intrinsic decomposition results (\cref{teaser}(c)), as well as NVS and relighting results (\cref{teaser}(b)). 
It also enables applications such as reflectance editing, relighting, and shading editing (\cref{teaser}(d)). Furthermore, our method is applicable across various datasets, including the object-only synthetic NeRF \cite{mildenhall2021nerf} dataset, the real object dataset \cite{gao2020deferred, zeng2023nrhints}, and the ReNe \cite{Toschi_2023_CVPR} dataset with real-world full scenes. Our contributions are summarized as follows:
\begin{itemize}
\item A novel intrinsic-aware NeRF model that integrates multiple light information, enabling applications such as NVS, lighting modification, and scene editing.
\item A method that separates intrinsic components by using supervision from generated pseudo intrinsic images. We introduce straightforward physics-based constraints to eliminate the need for statistical priors required by traditional approaches. Our method ensures robustness across various scene types.
\item Experimental results across three different datasets demonstrate our method's superior performance in intrinsic decomposition compared to existing state-of-the-art methods, showing advancements not only in synthetic object-only scenes but also in challenging real scenes with backgrounds and cast shadows.
\end{itemize}

\section{Related Works}
\noindent\textbf{{Intrinsic decomposition.}}
Intrinsic decomposition is a classical challenge in computer vision \cite{barrow1978recovering}, with much of the previous research focused on the 2D image \cite{dasPIENet, careagaIntrinsic, barron2014shape, li2018cgintrinsics}. A key difficulty in this area is the scarcity of real datasets, which need complicated and extensive annotation. This limitation has spurred interest in semi-supervised and unsupervised techniques \cite{BigTimeLi18,lettry2018unsupervised, liu2020unsupervised}. IntrinsicNeRF \cite{Ye2023IntrinsicNeRF} has been a pioneer in applying intrinsic decomposition to 3D neural rendering. Similar to previous unsupervised methods in 2D, it utilizes statistical priors, including chromaticity and semantic constraints, for guidance. However, these constraints do not accurately reflect physical principles and often fall short in complex scenarios. Our approach leans on 3D information and physical constraints (e.g., variations in illumination) to achieve superior results.

\noindent\textbf{{Relighting.}}
Relighting has recently garnered attention from various perspectives within the field \cite{einabadi2021deep}. Data-driven approaches have been explored, with research focusing on portrait scenes \cite{nestmeyer2020learning,sun2019single,zhou2019deep,pandey2021total, hou2022face} and extending to more complex scenarios \cite{murmann2019dataset, helou2020aim, puthussery2020wdrn, wang2020deep, el2021ntire}. Kocsis \etal \cite{kocsis2024lightit} have also investigated lighting control within diffusion models, enabling the generation of scenes under varying lighting conditions. Meanwhile, relighting has also received widespread attention within the field of neural rendering \cite{srinivasan2021nerv, gao2020deferred, zeng2023nrhints, Toschi_2023_CVPR}, achieving impressive relighting outcomes within individual scenes. Among them, Toschi \etal \cite{Toschi_2023_CVPR} proposed the ReNe dataset, which consists of images captured with various cameras and light poses under controlled lab conditions. Zeng \etal \cite{zeng2023nrhints} enhanced NeRF relighting with visibility and specular hints. Chang \etal \cite{chang2024fast} proposes a method of outdoor scene relighting, and they use locations and time to collect the direction of sunlight as input. 
However, the potential of using information from multiple lights for 3D scene understanding remains unexplored. 

\noindent\textbf{{Inverse rendering.}} Inverse rendering \cite{garces2022survey} is an alternative approach to recovering the fundamental properties of a scene, which aims at extracting the geometry materials, and lighting of a 3D scene. Recently, the study of inverse rendering methods based on NeRF has become a popular topic. NeRFactor \cite{zhang2021nerfactor} introduced a method to improve the geometric quality of NeRF and incorporated a data-driven BRDF prior. More methods have been developed to address the scenes with different light conditions, including fixed illumination conditions \cite{physg2021}, varying light sources \cite{boss2021nerd, boss2022-samurai}. Invrender \cite{zhang2022invrender} proposed a method for predicting indirect light. L-Tracing \cite{chen2022tracing} introduced an efficient algorithm for estimating visibility without training. SIRe-IR \cite{yang2023sireir} introduced a method for high-illuminance scenes, addressing the issue where previous methods struggled under prominent cast shadows. Liu \etal \cite{OpenIllumination} propose the OpenIllumination dataset with multi-illumination, which focuses on inverse rendering evaluation on real objects. However, inverse rendering methods are primarily based on individual objects and are challenging to extend to large, complex scenes, such as those with backgrounds.

\section{Method}

\begin{figure*}[ht!]
\centering
\includegraphics[width=1.0\linewidth]{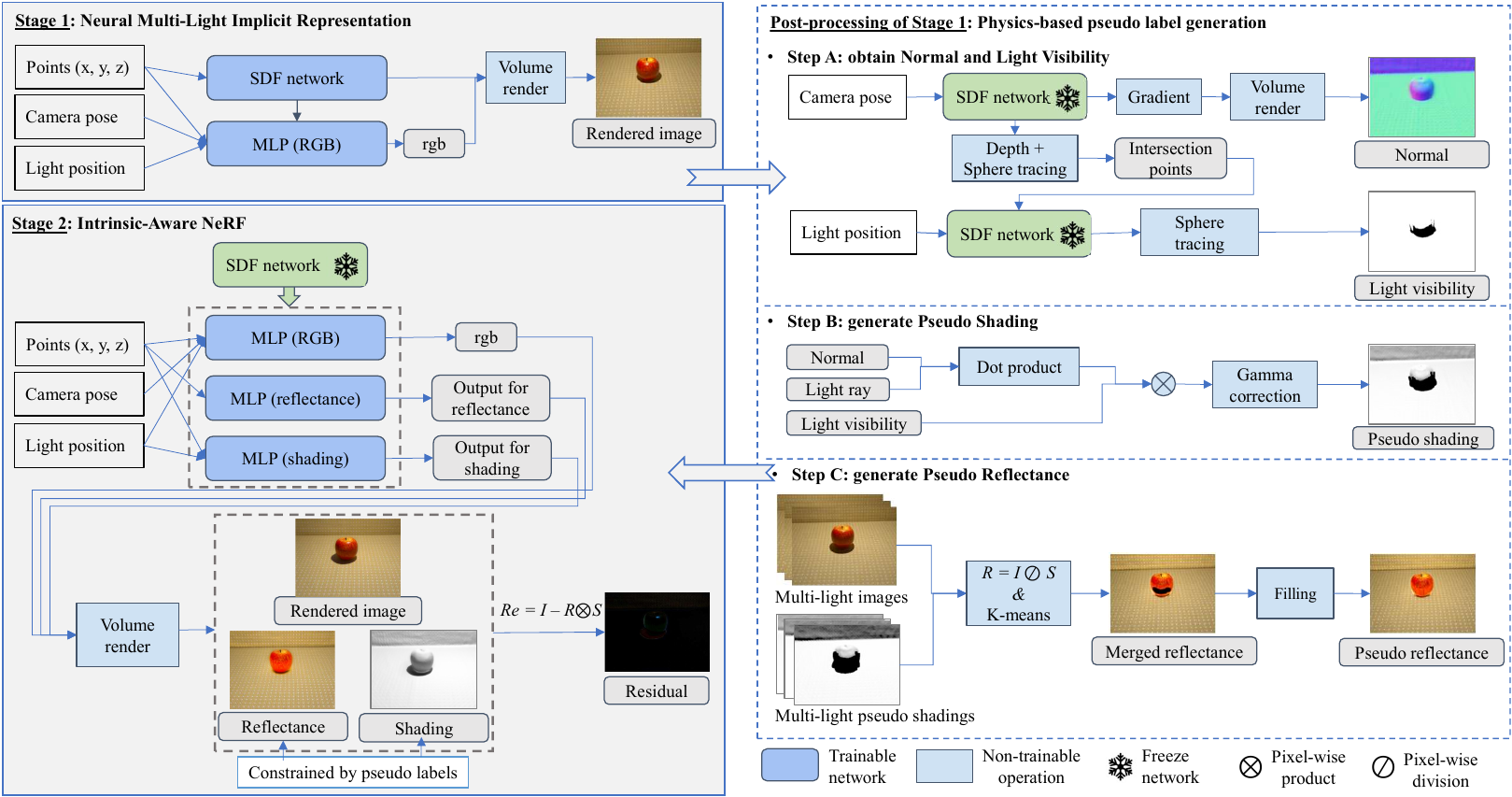}
\caption{\textbf{Illustration of the Framework.} In Stage 1, we introduce light position as input to extend NeRF for multi-light implicit representation (top left). Following Stage 1, three post-processing steps are applied to generate pseudo labels for reflectance and shading using the proposed physics-based pipeline (right). In Stage 2, we train the intrinsic-aware NeRF based on the model from Stage 1 and the pseudo labels from post-processing (bottom left).}
\label{Method Framework}
\vspace{-3mm}
\end{figure*}

We propose a two-stage method, with the overall framework illustrated in \cref{Method Framework}. 
In previous inverse rendering methods \cite{zhang2022invrender}, accounting for indirect illumination and shadow has been a critical challenge. Our strategy is to leverage multiple lighting conditions to make the corresponding information more accessible. 
In the first stage, we train our model to represent scenes under varying camera positions and lighting conditions, enabling NVS and lighting modification. 
In the subsequent post-processing, we use the model from Stage 1 to generate pseudo intrinsic images of reflectance and shading for each image. 
Specifically, pseudo shadings are derived from normals and light visibility, while pseudo reflectance is obtained by dividing the image by the shading.
Leveraging multiple lighting conditions allows us to effectively gather more detailed information about the scene, thus resulting in high-fidelity pseudo labels.
In Stage 2, we retrain the network with added modules for predicting reflectance and shading, restricting light information input solely to the shading module. Pseudo images guide the separation of intrinsic components in this phase.

\subsection{Preliminaries}
Some traditional methods decompose images into reflectance and shading \cite{barrow1978recovering,careagaIntrinsic,fan2018revisiting}, primarily targeting diffuse components. Since the real world contains many non-diffuse effects, recent approaches\cite{garces2022survey, Ye2023IntrinsicNeRF} add a residual term to account for discrepancies. 
We follow this setup and model intrinsic decomposition as follows:
\begin{equation}
I(i, j)=R(i, j) \otimes S(i, j) + Re(i, j)
\label{eq: intrinsic decomposition}
\end{equation}
\noindent where $R$, $S$ and $Re$ denote Reflectance, Shading and Residual, respectively. 
Additionally, following the Lambert’s cosine law, a shading can be computed using the following formula:
\begin{equation}
S = \Vec{N} \cdot \Vec{L}
\label{Lambertia law}
\end{equation}
where $\Vec{N}$ is the normal ray and $\Vec{L}$ is the light ray. 
This physical illumination model guides our following pseudo shading generation and intrinsic-aware NeRF training.

\subsection{Stage 1: Neural Multi-Light Implicit Representation}
We follow the structure of Neuralangelo \cite{li2023neuralangelo}, which achieves promising results in both small and large scenes. It uses 3D hashing encoding combined with Signed Distance Function (SDF) \cite{mueller2022instant} to represent the implicit geometry, and then employs an MLP to model the color information. 
The original Neuralangelo does not support light position as an input, so, besides the original MLP input, we incorporate the light position encoded with spherical encoding as an additional input. 
We illustrate Stage 1 in \cref{Method Framework} (top-left), with formulas as follows:
\begin{equation}
sdf = f(\mathbf{x}), \quad \mathbf{c} = \textrm{MLP}_{color}(\mathbf{x}, \mathbf{n}, \mathbf{feat},\mathbf{d}, \mathbf{l})
\label{eq: stage 1}
\end{equation}
where $f(\cdot)$ is the geometry network that predicts SDF and $\textrm{MLP}_{color}(\cdot)$ is the color network. $\mathbf{x}$ is the spatial position, $\mathbf{n}$ and $\mathbf{feat}$ are the normal and the features from the SDF network, $\mathbf{d}$ is the view direction, and $\mathbf{l}$ is the light position.
Following \cite{li2023neuralangelo}, the loss for Stage 1 is:
\begin{equation}
\mathcal{L}_{S1} = w_{\text{rgb}}\mathcal{L}_{\mathrm{rgb}} + w_{\text{eik}} \mathcal{L}_{\text{eik}} + w_{\text{curv}} \mathcal{L}_{\text{curv}}
\end{equation}
where $\mathcal{L}_{\mathrm{rgb}}$ is the loss of the rendered image, $\mathcal{L}_{\text{eik}}$ represents the Eikonal loss \cite{gropp2020implicit}, and $\mathcal{L}_{\text{curv}}$ is the curvature loss \cite{li2023neuralangelo}. The terms $w_{\text{eik}}$ and $w_{\text{curv}}$ are the corresponding weights.

\subsection{Post-processing: Physics-based Pseudo Label Generation}
Here we propose a post-processing that generate pseudo labels for reflectance and shading in three steps, as illustrated in \cref{Method Framework} (right).
It starts with generating pseudo shading from the normal and the light visibility. 
We then generate pseudo reflectance using multi-light shadings and images.

\noindent \textbf{Step A.} 
We derive the normal from the gradient of the SDF network. The geometry network also provides depth information which is used to estimate the intersection points in conjunction with sphere tracing \cite{chen2022tracing}. 
Light visibility, denoted as $V$, which indicates whether a point is directly illuminated, is obtained by sphere tracing based on the light position and intersection points.

\noindent \textbf{Step B.} 
Since the \cref{Lambertia law} does not consider occlusion and other effects, the generation of pseudo-shading in our implementation follows the formula:
\begin{equation}
S^* = (\max(\Vec{N} \cdot \Vec{L}, 0) \otimes V)^\gamma
\end{equation}
where $(\cdot)^\gamma$ represents gamma correction, and $\otimes$ denotes pixel-wise product. 
This gamma correction is essential to adapt to the nonlinear representation of digital images. All image sensors introduce a gamma correction to accommodate the light integration to the nonlinear perception of brightness in the human eye. 
Since our pseudo-shading is directly created from the 3D world, we need to introduce the sensor representation.

\noindent \textbf{Step C.} 
This step entails inferring high-fidelity pseudo reflectance from multiple pseudo shading. We use the equation $R=I\oslash S$ as a simplified version of \cref{eq: intrinsic decomposition} ignoring at this point the residual component that mainly entangles specular and undirected light components. 

\begin{figure*}[t]
\centering
\includegraphics[width=0.8\linewidth]{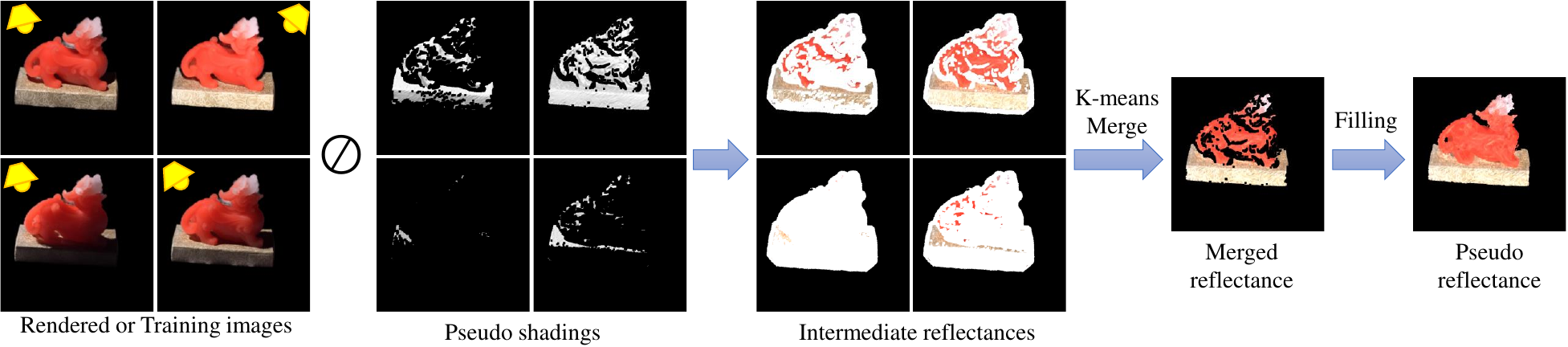}
\vspace{-2mm}
\caption{Illustration of the pseudo reflectance generation process in the post-processing.}
\label{fig: flowchart of the generation of pseudo reflectance}
\vspace{-3mm}
\end{figure*}

Our method in this step leverages the trained model from Stage 1 to generate multiple images under different direct light conditions, each accompanied by its corresponding pseudo shading. The flowchart is shown at the right bottom in \cref{Method Framework}, and \cref{fig: flowchart of the generation of pseudo reflectance} illustrates the images during the calculation. 
By calculating $R=I\oslash S$, we can compile the reflectance information for every pixel from multiple light conditions while diminishing the influence of the entangled indirect light. 
To generate a unique reflectance from multiple pseudo shading, we use the K-means algorithm \cite{scikit-learn} at the pixel level, incorporating the weights of each pseudo shading to select the most probable pseudo reflectance. This approach allows us to achieve a merged reflectance under varied lighting conditions.

However, some regions within the merged reflectance may appear as holes due to the absence of direct illumination in all lighting conditions. We address these areas with a filling strategy. We compute a weighted distance between hole pixels and non-hole pixels, considering their spatial distance in the image, the angular difference of their normals, and the color difference in the RGB image. Then, we assign the color of the nearest non-hole pixel, based on this weighted distance, to the hole pixel. In fact, our filling strategy is a rough supplementary approach with minimal impact on the overall results. Under the multi-light setting, the hole regions are typically small. Additionally, we reduced the weight of the hole regions during subsequent training.

After our proposed three-step post-processing, we achieve the final pseudo reflectance, as shown in \cref{fig: flowchart of the generation of pseudo reflectance}. Our pseudo reflectance offers a straightforward basis for disentanglement and has proven to be more reliable in guiding the following training stage compared to the previous statistical priors.
Additionally, we compute weight maps $W_R$ and $W_S$ for both pseudo reflectance and pseudo shading based on the edges of pseudo shading and visibility. Areas with higher pseudo shading values, or those further from visibility edges (where visibility calculations may be prone to errors), exhibit greater credibility in their pseudo labels; conversely, areas closer to visibility edges or with lower pseudo shading values are deemed less reliable.

\subsection{Stage 2: Intrinsic-Aware NeRF}
As illustrated in \cref{Method Framework} (bottom-left), we use high-fidelity pseudo intrinsic images to guide the intrinsic decomposition learning. 
Expanding the model from Stage 1, we add two extra MLPs dedicated to generating reflectance and shading outputs, while the geometry network is frozen. 
Compared to the $\textrm{MLP}_{color}$ in \cref{eq: stage 1}, these two MLPs receive different inputs. 
Since reflectance is independent of lighting, the $\textrm{MLP}_{reflectance}$ does not take the light position as input. 
Additionally, both reflectance and shading are diffuse components, and a diffuse reflecting surface exhibits Lambertian reflection, indicating that it maintains equal luminance when observed from any direction. Therefore, view poses are also excluded from both of their inputs. 
The formulas are as follows:
\begin{equation}
    \begin{split}
& \mathbf{r} = \textrm{MLP}_{reflectance}(\mathbf{x}, \mathbf{n}, \mathbf{feat})
\\
& \mathbf{s} = \textrm{MLP}_{shading}(\mathbf{x}, \mathbf{n}, \mathbf{feat}, \mathbf{l})
    \end{split}
\end{equation}
where $\mathbf{x}$ is the spatial position, $\mathbf{n}$ and $\mathbf{feat}$ are the normal and the features from the SDF network, and $\mathbf{l}$ is the light position.

After volume rendering, we obtain RGB images, along with reflectance and shading. Subsequently, the residual is derived from \cref{eq: intrinsic decomposition}. 
During training, the pseudo labels impose constraints on reflectance and shading: 
\begin{equation}
L_{intrinsic}=W_R \cdot \|\hat{R}-R^*\|_1 + W_S \cdot \|\hat{S}-S^*\|_1
\label{eq: intrinsic loss}
\end{equation}
where $\hat{R}$ and $\hat{S}$ represent the predicted reflectance and shading, respectively, and $R^*$ and $S^*$ are their corresponding pseudo labels. 
$W_R$ and $W_S$ represent weight maps for reflectance and shading, derived during pseudo label generation.
As demonstrated in \cite{Ye2023IntrinsicNeRF}, the diffuse components dominate the scene, so it is crucial to prevent the training from converging to undesirable local minima (\eg $R=0, S=0, Re=I$). 
Therefore, we introduce a regularization term for $Re$ to ensure that the image is primarily recovered through $R$ and $S$: $L_{reg}=\|\hat{Re}\|_1$.

Finally, the Stage 2 loss is the weighted sum of:
\begin{equation}
\mathcal{L}_{S2} = w_{\text{rgb}}\mathcal{L}_{\mathrm{rgb}} + w_{\text{intrinsic}}L_{intrinsic} + w_{\text{reg}}L_{reg}
\end{equation}
where $w_{\text{rgb}}$, $w_{\text{intrinsic}}$ and $w_{\text{reg}}$ are the corresponding weights.

\section{Experiments}

\begin{table*}[!t]
\centering
\resizebox{0.8\linewidth}{!}{
\begin{tabular}{cccccccccc}
\hline
\multirow{2}{*}{Method} & \multirow{2}{*}{Light Setting} & \multicolumn{4}{c}{Reflectance}  & \multicolumn{4}{c}{Shading}      \\
                        &                          & PSNR$\uparrow$  & SSIM$\uparrow$   & LPIPS$\downarrow$                & MSE $\downarrow$    & PSNR$\uparrow$  & SSIM$\uparrow$   & LPIPS$\downarrow$                & MSE $\downarrow$    \\
                        \hline
InvRender     & Single   & 16.59 & 0.8228 & 0.1807 & 0.0271 & ---   & ---    & ---    & ---    \\
TensoIR       & Single   & 18.50 & 0.8518 & 0.1544 & 0.0213 & ---   & ---    & ---    & ---    \\
IntrinsicNeRF & Single   & 18.02 & 0.8353 & 0.2142 & 0.0226 & 19.02 & 0.8660 & 0.1476 & 0.0168 \\
Ours          & Single   & 22.44 & 0.9012 & 0.0950 & 0.0072 & \textbf{24.39} & \textbf{0.9188} & \textbf{0.0843} & \textbf{0.0049} \\ \cline{7-10} 
TensoIR       & Multiple & 22.70 & 0.8800 & 0.1450 & 0.0087 & ---   & ---    & ---    & ---    \\
Ours          & Multiple & \underline{24.40} & \underline{0.9357} & \underline{0.0582} & \underline{0.0051} & 23.44 & 0.9225 & 0.0798 & 0.0055 \\  \cline{7-10} 
PIE-Net       & Random   & 19.59 & 0.8708 & 0.1298 & 0.0153 & 20.03 & 0.8868 & 0.1653 & 0.0133 \\
Ordinal       & Random   & 18.14 & 0.8716 & 0.1251 & 0.0191 & 16.80 & 0.8717 & 0.1679 & 0.0293 \\
Ours          & Random   & \textbf{25.48} & \textbf{0.9420} & \textbf{0.0515} & \textbf{0.0041} & \textbf{22.90} & \textbf{0.9024} & \textbf{0.1049} & \textbf{0.0072} \\
\hline
\end{tabular}
}
\caption{Quantitative results of the intrinsic decomposition on the \textbf{Synthetic Dataset}. We compare different methods under three light settings: single, multiple, and random. Our method outperforms other methods in all settings. Best results are marked in \textbf{bold}, second best results are \underline{underlined}.}
\label{tab: syn iid}
\vspace{-3mm}
\end{table*}

We quantitatively and qualitatively validate our method and compare it with other approaches, including traditional learning-based intrinsic decomposition methods and neural rendering methods. 
The comparison encompasses intrinsic decomposition and NVS with relighting.
For data-driven methods, we select PIE-Net\cite{dasPIENet} and Careaga \etal\cite{careagaIntrinsic} (hereafter referred to as Ordinal). 
For NeRF-related methods, we choose InvRender\cite{zhang2022invrender}, TensoIR\cite{Jin2023TensoIR}, and IntrinsicNeRF\cite{Ye2023IntrinsicNeRF}. Among these, InvRender and TensoIR are inverse rendering methods, while IntrinsicNeRF is an intrinsic decomposition method. Additionally, we select NRHints\cite{zeng2023nrhints}, a method focused on relighting, to validate our relighting performance.

\subsection{Datasets}
Our experiments are conducted on three datasets, each containing four scenes.

\noindent \textbf{Synthetic Dataset.}
It contains synthetic data based on Blender, inspired by the synthetic dataset designed by NRHints\cite{zeng2023nrhints} for relighting. Some scenes in this dataset are derived from NeRF\cite{mildenhall2021nerf}. Since the original dataset does not include ground truth (GT) for intrinsic decomposition, we re-render the data in Blender and export these quantities. 
Each scene comprises 500 images for training, 100 for validation, and 100 for testing, including intrinsic components for each image. The selection of light positions and camera poses follows the setup described by Zeng et al. \cite{zeng2023nrhints}, with both distributed across a hemispherical space above the scene. The light position and camera pose for each image are randomly and independently selected.

\noindent \textbf{Real Object Dataset.}
It includes real objects from the object relighting dataset collected by \cite{gao2020deferred, zeng2023nrhints}, featuring various viewpoints and lighting conditions. The training set size varies from 500 to 2000 images per scene.

\noindent \textbf{ReNe Dataset.}
ReNe dataset\cite{Toschi_2023_CVPR}, unlike most datasets used in previous inverse rendering research, contains complete real-world scenes with backgrounds. 
Notably, the camera poses and light positions in this dataset are concentrated in a specific direction rather than evenly distributed across 360 degrees, which poses challenges for both reconstruction and intrinsic decomposition.
This real dataset features 2000 images across scenes, captured from 50 viewpoints under 40 lighting conditions, with lighting and camera poses grid-sampled. 
Following the dataset split, we use 1628 images (44 camera poses $\times$ 37 light positions) for training. 
Since the test set is not publicly available, we use the validation set for inference.

\subsection{Lighting and Camera Views Settings}

In this section, we further clarify the lighting and camera view setups in our experiments.

\noindent \textbf{Grid-sampled or non-grid-sampled.}
In the ReNe dataset, camera views and light positions are grid-sampled. In contrast, the Synthetic and Real Object Datasets use independently sampled views and lights. We denote this as non-grid-sampled. This difference does not affect Stage 1 training but influences the post-processing of Stage 1 (\cref{Method Framework}), where we merge results to generate pseudo reflectance.For grid-sampled setups, all light positions are used. For non-grid-sampled setups, because the number of possible light positions is large, four light positions are randomly selected, which our experiments indicate is sufficient.

\noindent \textbf{Additional Lighting Setups for Better Comparison.} Previous methods often struggle with varying lighting conditions. InvRender \cite{zhang2022invrender} and IntrinsicNeRF \cite{Ye2023IntrinsicNeRF} operate under single lighting conditions, while TensoIR \cite{Jin2023TensoIR} supports multiple lights but recommends around four, which cannot handle hundreds of light positions. We introduce two additional settings to better compare intrinsic decomposition methods and examine the impact of different light positions: \textbf{single light} (one fixed position) and \textbf{multiple lights} (four fixed positions). TensoIR occasionally failed with four lights, so we used three instead in this case.

For the Synthetic Dataset, the original setting is referred to as \textbf{random lights}. We re-render scenes in Blender for the single and multiple light tracks, maintaining the same image counts and random seed for camera views. For the ReNe Dataset, the original setting is denoted as \textbf{all lights}. We selected 1 light condition as the single light track and 4 light conditions as the multiple lights track, thereby the training sets respectively contain $44\times1$ and $44\times4$ images. Notably, the single light setup serves as an extreme condition, diverging from our intent of utilizing multi-light information. This track was designed for fair comparison with other methods and to highlight the importance of multi-light information in intrinsic decomposition.

\subsection{Implementation Details}
Our model's hyperparameters include a batch size of 2048 and each stage is trained for 500k iterations. 
We implement the model in PyTorch and use the AdamW \cite{loshchilov2018decoupled} optimizer with a learning rate of $1e^{-3}$ for optimization. 
The experiments can be conducted on a single Nvidia RTX 3090. The training time of Stage 1 follows the training time of Neuralangelo \cite{li2023neuralangelo}, and Stage 2 takes 19 hours. The weights of losses, $w_{\text{eik}}$, $w_{\text{curv}}$, $w_{\text{intrinsic}}$, $w_{\text{reg}}$ are set to $0.1$, $5e^{-4}$, $1.0$, and $1.0$, respectively. 
To evaluate the comparison between predicted images and GT, we employ the following metrics: Peak Signal-to-Noise Ratio (PSNR), Structural Similarity Index (SSIM) \cite{wang2004image}, Learned Perceptual Image Patch Similarity (LPIPS) \cite{zhang2018unreasonable} and Mean Squared Error (MSE).

\subsection{Results on the Synthetic Dataset}

\begin{figure*}[!ht]
\centering
\includegraphics[width=\linewidth]{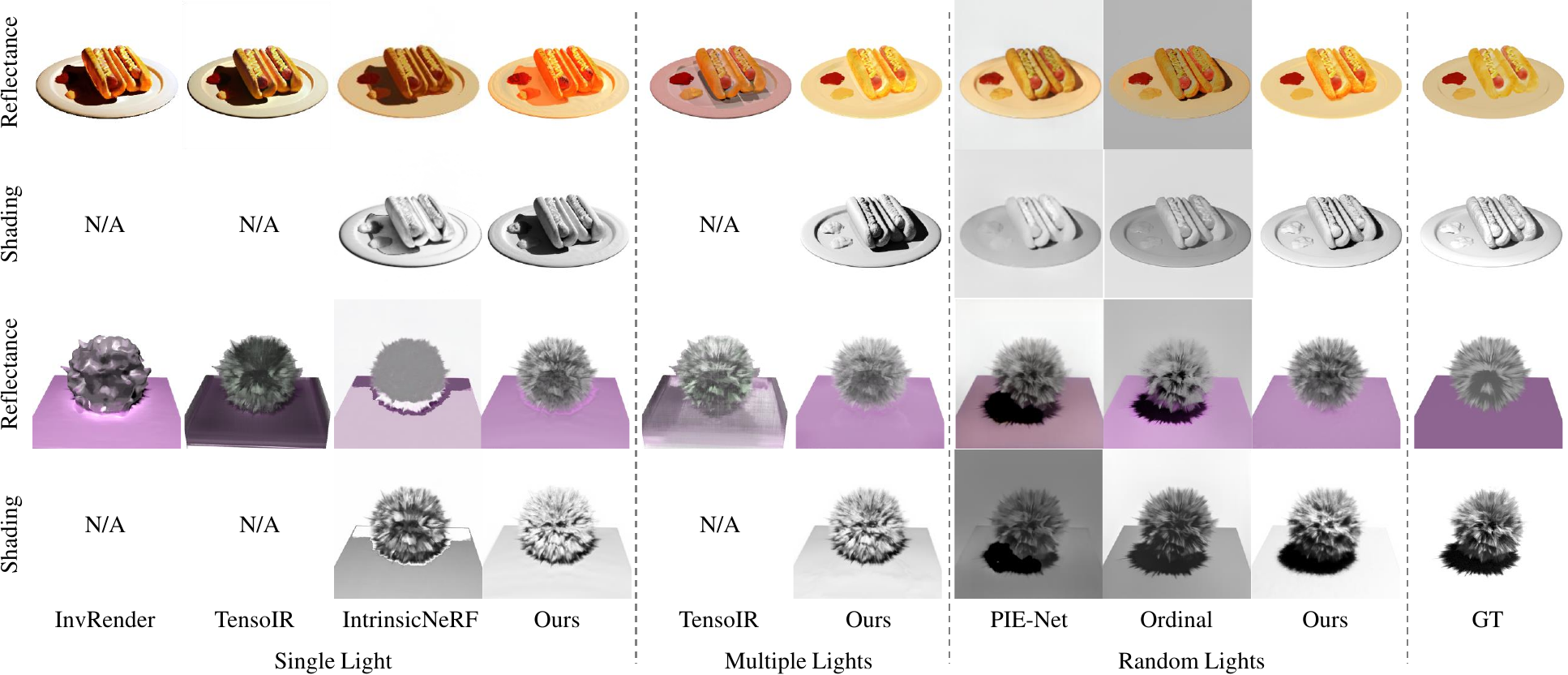}
\vspace{-6mm}
\caption{Qualitative Results on the \textbf{Synthetic Dataset} with all settings. The same GT reflectance applies across all settings, but GT shading differs due to varying light positions. Here for brevity, only the shading under the Random Lights setting is shown. Compared to other methods, our approach predicts the best reflectance and effectively handles cast shadows.}
\label{fig: syntheic}
\vspace{-3mm}
\end{figure*}

\noindent\textbf{Comparison with SOTA.} 
\cref{tab: syn iid} presents the quantitative comparison of intrinsic decomposition on all methods and light settings. 
Since the sequence of camera poses is the same across all settings when generating the dataset, the reflectance GT remains consistent, allowing for direct comparison. 
Our method outperforms existing methods in predicting reflectance for each light setting, with the best results under the random setting.
Due to the different shading GTs, the comparisons are conducted separately. Our method also achieves the best results across all settings.
A qualitative comparison is shown in \cref{fig: syntheic}. The single light setting demonstrates that under challenging lighting conditions, such as those with pronounced cast shadows, all methods struggle to achieve consistently good results. This indicates that using multi-light information is a practical approach. However, our method still achieves the most meaningful results in the single light setting of the FurBall scene. In the other two settings, our approach also demonstrates significantly better performance, producing high-quality reflectance and shading results, effectively handling cast shadows.

\noindent\textbf{Comparison across different light settings.} 
As shown in \cref{tab: syn iid} and \cref{fig: syntheic}, both TensoIR and our method exhibit improved performance as the number of light positions increases. This observation aligns with the findings mentioned in the TensoIR \cite{Jin2023TensoIR}. 
Additionally, the results in \cref{fig: syntheic} demonstrate that our method's performance under the multiple light setting (4 lights) is very close to that of the random light setting.

\noindent\textbf{NVS and relighting analysis.}
\cref{tab: syn nvs relighting} shows the quantitative results of NVS and relighting. Since the lights are fixed and known under the Single and Multiple light settings, the results pertain only to NVS, where our method achieves the best performance. Under the Random setting, the results reflect both NVS and relighting, where our method shows comparable performance to NRHints \cite{zeng2023nrhints}.

\begin{table}[!ht]
\centering
\resizebox{0.95\linewidth}{!}{
\begin{tabular}{cccccc}
\hline
        & Light Setting  & PSNR$\uparrow$  & SSIM$\uparrow$   & LPIPS$\downarrow$                & MSE $\downarrow$                  \\
        \hline
InvRender     & Single   & 23.99 & 0.8760 & 0.1109 & 0.0051 \\
TensoIR       & Single   & 35.00 & 0.9761 & 0.0343 & 0.0019 \\
IntrinsicNeRF & Single   & 34.53 & 0.9794 & 0.0137 & 0.0005 \\
Ours          & Single   & \textbf{37.51} & \textbf{0.9878} & \textbf{0.0099} & \textbf{0.0003} \\
\hline
TensoIR       & Multiple & 33.19 & 0.9645 & 0.0503 & 0.0022 \\
Ours          & Multiple & \textbf{36.12} & \textbf{0.9836} & \textbf{0.0135} & \textbf{0.0003} \\
\hline
NRHints       & Random   & \textbf{32.79} & \textbf{0.9674} & 0.0369 & \textbf{0.0007} \\
Ours          & Random   & 31.20 & 0.9619 & \textbf{0.0354} & 0.0010       \\
\hline
\end{tabular}
}
\caption{Quantitative results of the NVS and relighting on the \textbf{Synthetic Dataset}.}
\label{tab: syn nvs relighting}
\vspace{-3mm}
\end{table}

\subsection{Results on the Real Object Dataset}

\begin{figure*}[!ht]
\centering
\includegraphics[width=\linewidth]{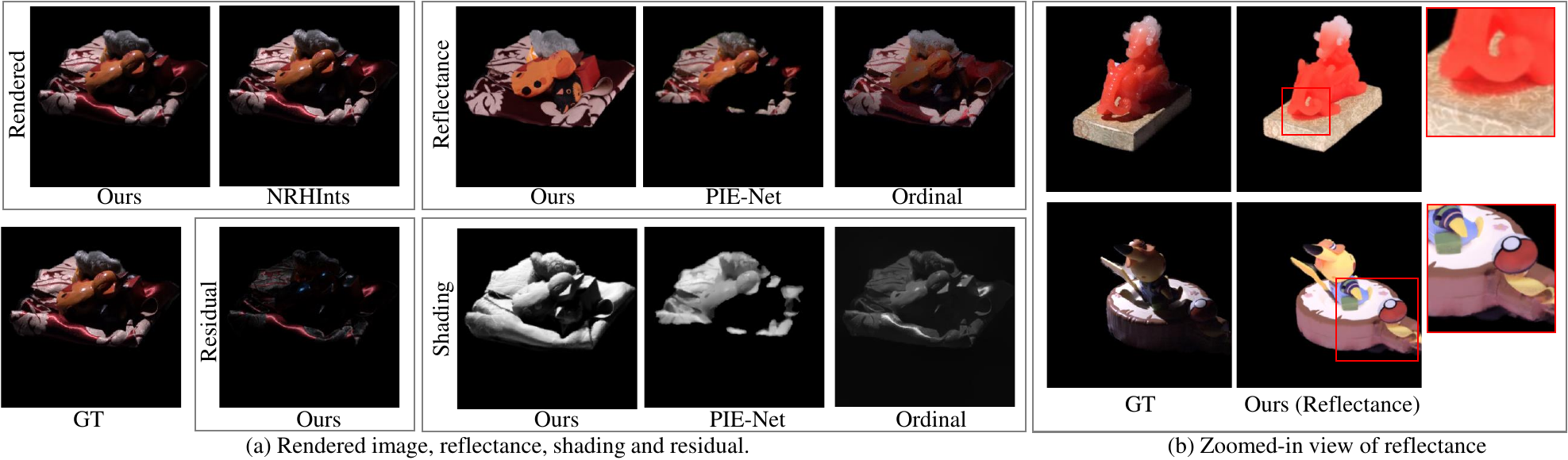}
\vspace{-3mm}
\caption{Qualitative Results on the \textbf{Real Object Dataset}. (a) Our method compared with NRHints for the rendered image, and PIE-Net and Ordinal for intrinsic decomposition. (b) Our reflectance estimation for two different scenes, with zoomed-in views on the object hole and cast shadow area.}
\label{fig: real}
\vspace{-1mm}
\end{figure*}

Results on the Real Object Dataset allow us to confirm a quite accurate qualitative performance of our intrinsic decomposition. In \cref{fig: real}(a) we show a comparison of our method versus SOTA. Our RGB-rendered images are similar to NRHInts \cite{zeng2023nrhints} and our estimations for Reflectance and Shading can be compared to PIE-Net \cite{dasPIENet} and Ordinal \cite{careagaIntrinsic}. In particular, we want to highlight our estimation of reflectance that does not present any remaining shading effect with respect to the rest of the methods, restoring the vibrant colors of the objects and achieving significantly better results. In \cref{fig: real}(b) we add two more examples where we zoom out two details. At the top is a hole through which our method estimates the texture of the support surface, and at the bottom, it gets a proper reflectance from a strong cast shadow. Our shading presents a better disentangling from reflectance than the rest of the methods. Although we can show a clear overall improvement, some limitations can still be observed from some difficult areas that remain hidden from any camera point of view and light source, as we can see in the unremoved shadow we have under the object's tail of the top zoomed window.

\subsection{Results on the ReNe Dataset}

\begin{figure*}[!ht]
\centering
\includegraphics[width=\linewidth]{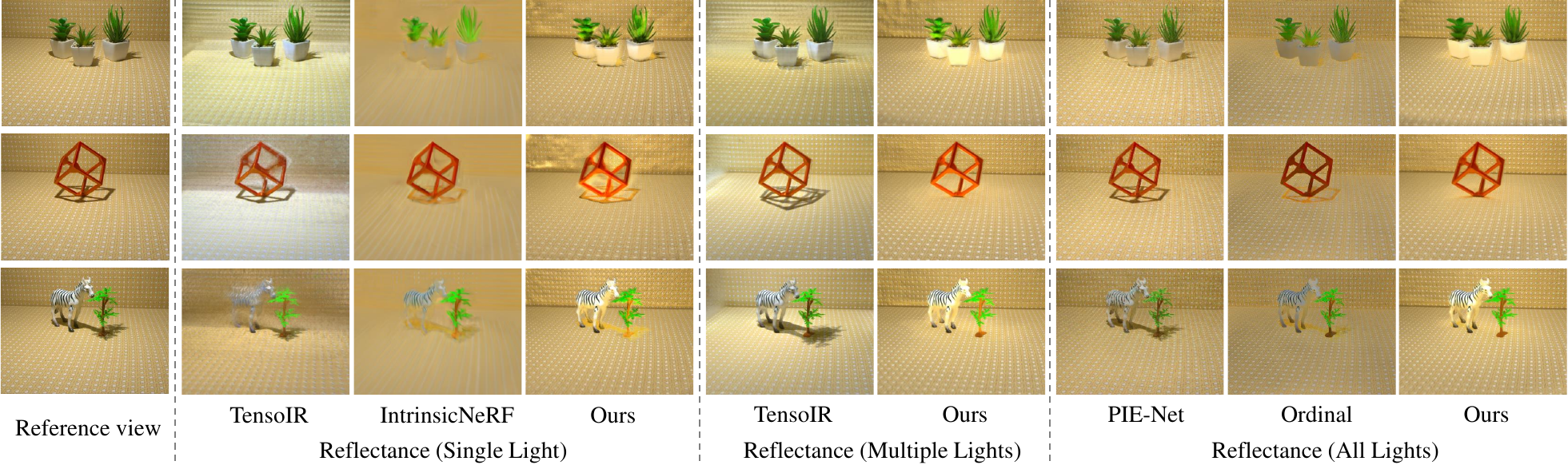}
\vspace{-3mm}
\caption{Qualitative Results on the \textbf{ReNe Dataset}. We show the reflectance estimation for a reference view across all settings.}
\label{fig: rene}
\vspace{-5mm}
\end{figure*}

The ReNe dataset presents a significant challenge for previous methods due to its full scenes with backgrounds and camera poses that are mostly concentrated in a specific area rather than being distributed 360 degrees around the scene.
The quantitative results of NVS in \cref{tab: rene} also demonstrate that our method gets the best results in all settings across all metrics.

\setlength{\tabcolsep}{3pt}
\begin{table}[!ht]
\vspace{-1mm}
\centering
\resizebox{0.95\linewidth}{!}{
\begin{tabular}{cccccc}
\hline
              & Light Setting  & PSNR$\uparrow$  & SSIM$\uparrow$   & LPIPS$\downarrow$                & MSE $\downarrow$ \\
              \hline
TensoIR       & Single   & 24.57 & 0.6063 & 0.1912 & 0.0040 \\
IntrinsicNeRF & Single   & 24.10 & 0.4009 & 0.5392 & 0.0039 \\
Ours          & Single   & \textbf{27.63} & \textbf{0.7210} & \textbf{0.1136} & \textbf{0.0019} \\
\hline
TensoIR       & Multiple & 26.44 & 0.6743 & 0.1572 & 0.0028 \\
Ours          & Multiple & \textbf{27.71} & \textbf{0.7377} & \textbf{0.1078} & \textbf{0.0018} \\
\hline
Ours          & All      & 27.48 & 0.7300 & 0.1113 & 0.0019 \\
\hline
\end{tabular}
}
\caption{Quantitative results of the NVS and relighting on the \textbf{ReNe Dataset}.}
\label{tab: rene}
\end{table}

The results from the single light setting (left of \cref{fig: rene}) show that all methods struggle to achieve good performance; however, our method produces meaningful results in the first and third scenes. 
The other neural rendering methods, TensoIR\cite{Jin2023TensoIR} and IntrinsicNeRF \cite{Ye2023IntrinsicNeRF}, fail to achieve correct decomposition, primarily attributed to the failure in distinguishing intrinsic components and also the difficulty in scene reconstruction. 
In the multiple light setting(middle of \cref{fig: rene}), TensoIR incorrectly leaks multiple shading components into the reflectance. In contrast, our method successfully combines multiple lighting conditions to produce clean reflectance. 
In the all lights setting, our method achieves superior results, outperforming both PIE-Net \cite{dasPIENet} and Ordinal \cite{careagaIntrinsic}, achieving sharp texture edges, vibrant colors, accurate shadow elimination, and precise background reconstruction. For more results, please refer to the supplementary material.

\section{Conclusion}
We propose MLI-NeRF, a Multi-light Intrinsic-aware Neural Radiance Field. MLI-NeRF generates pseudo intrinsic images for scenes under different lighting conditions, enabling the learning of intrinsic decomposition without intrinsic GT. Our experiments demonstrate that our method achieves excellent performance across different types of scenes. Our approach relies on fundamental physical principles, highlighting its potential applicability in more complex scenes. Our method enables the simultaneous synthesis of novel views, relighting, and intrinsic decomposition, providing a versatile tool for various editing applications, such as reflectance and shading modification.

\noindent \textbf{Limitations and Future Work.} 
One limitation of our method is the need to know the light positions. As noted in our related works, many methods have proposed ways to obtain light pose information when capturing custom data. These include recording the camera pose fixed with the light source\cite{ling2022shadowneus,zeng2023nrhints} and using the time of day to determine the sun's angle in outdoor settings\cite{chang2024fast}.
Another limitation is the computational efficiency: training the Stage 1 model based on Neuralangelo \cite{li2023neuralangelo} takes around 40 hours, with Stage 2 retraining adding another 19 hours. In fact, our method is applicable to different similar baselines, and we plan to migrate to a more efficient base architecture in the future.

\section*{Acknowledgments}
This work was supported by Grant PID2021-128178OB-I00 funded by MCIN/AEI/10.13039/501100011033 and by ERDF "A way of making Europe". It was also supported by the Departament de Recerca i Universitats from Generalitat de Catalunya, reference 2021SGR01499.  Yixiong Yang is supported by China Scholarship Council.

{
    \small
    \bibliographystyle{ieeenat_fullname}
    \bibliography{main}

\begin{thebibliography}{50}
\providecommand{\natexlab}[1]{#1}
\providecommand{\url}[1]{\texttt{#1}}
\expandafter\ifx\csname urlstyle\endcsname\relax
  \providecommand{\doi}[1]{doi: #1}\else
  \providecommand{\doi}{doi: \begingroup \urlstyle{rm}\Url}\fi

\bibitem[Barron and Malik(2014)]{barron2014shape}
Jonathan~T Barron and Jitendra Malik.
\newblock Shape, illumination, and reflectance from shading.
\newblock \emph{IEEE transactions on pattern analysis and machine intelligence}, 37\penalty0 (8):\penalty0 1670--1687, 2014.

\bibitem[Barrow et~al.(1978)Barrow, Tenenbaum, Hanson, and Riseman]{barrow1978recovering}
Harry Barrow, J Tenenbaum, A Hanson, and E Riseman.
\newblock Recovering intrinsic scene characteristics.
\newblock \emph{Comput. Vis. Syst}, 2\penalty0 (3-26):\penalty0 2, 1978.

\bibitem[Boss et~al.(2021)Boss, Braun, Jampani, Barron, Liu, and Lensch]{boss2021nerd}
Mark Boss, Raphael Braun, Varun Jampani, Jonathan~T Barron, Ce Liu, and Hendrik Lensch.
\newblock Nerd: Neural reflectance decomposition from image collections.
\newblock In \emph{Proceedings of the IEEE/CVF International Conference on Computer Vision}, pages 12684--12694, 2021.

\bibitem[Boss et~al.(2022)Boss, Engelhardt, Kar, Li, Sun, Barron, Lensch, and Jampani]{boss2022-samurai}
Mark Boss, Andreas Engelhardt, Abhishek Kar, Yuanzhen Li, Deqing Sun, Jonathan~T. Barron, Hendrik~P.A. Lensch, and Varun Jampani.
\newblock {SAMURAI}: {S}hape {A}nd {M}aterial from {U}nconstrained {R}eal-world {A}rbitrary {I}mage collections.
\newblock In \emph{Advances in Neural Information Processing Systems (NeurIPS)}, 2022.

\bibitem[Burley and Studios(2012)]{burley2012physically}
Brent Burley and Walt Disney~Animation Studios.
\newblock Physically-based shading at disney.
\newblock In \emph{Acm Siggraph}, pages 1--7. vol. 2012, 2012.

\bibitem[Careaga and Aksoy(2023)]{careagaIntrinsic}
Chris Careaga and Ya\u{g}{\i}z Aksoy.
\newblock Intrinsic image decomposition via ordinal shading.
\newblock \emph{ACM Trans. Graph.}, 2023.

\bibitem[Chang et~al.(2024)Chang, Kim, Seo, Yi, and Kwak]{chang2024fast}
Yeonjin Chang, Yearim Kim, Seunghyeon Seo, Jung Yi, and Nojun Kwak.
\newblock Fast sun-aligned outdoor scene relighting based on tensorf.
\newblock In \emph{Proceedings of the IEEE/CVF Winter Conference on Applications of Computer Vision}, pages 3626--3636, 2024.

\bibitem[Chen et~al.(2022)Chen, Ding, Guo, Wang, Li, Xiao, Wu, and Song]{chen2022tracing}
Ziyu Chen, Chenjing Ding, Jianfei Guo, Dongliang Wang, Yikang Li, Xuan Xiao, Wei Wu, and Li Song.
\newblock L-tracing: Fast light visibility estimation on neural surfaces by sphere tracing.
\newblock In \emph{Proceedings of the European Conference on Computer Vision (ECCV)}, 2022.

\bibitem[Das et~al.(2022)Das, Karaoglu, and Gevers]{dasPIENet}
Partha Das, Sezer Karaoglu, and Theo Gevers.
\newblock Pie-net: Photometric invariant edge guided network for intrinsic image decomposition.
\newblock In \emph{IEEE Conference on Computer Vision and Pattern Recognition, (CVPR)}, 2022.

\bibitem[Einabadi et~al.(2021)Einabadi, Guillemaut, and Hilton]{einabadi2021deep}
Farshad Einabadi, Jean-Yves Guillemaut, and Adrian Hilton.
\newblock Deep neural models for illumination estimation and relighting: A survey.
\newblock In \emph{Computer Graphics Forum}, pages 315--331. Wiley Online Library, 2021.

\bibitem[El~Helou et~al.(2021)El~Helou, Zhou, Susstrunk, and Timofte]{el2021ntire}
Majed El~Helou, Ruofan Zhou, Sabine Susstrunk, and Radu Timofte.
\newblock Ntire 2021 depth guided image relighting challenge.
\newblock In \emph{Proceedings of the IEEE/CVF Conference on Computer Vision and Pattern Recognition}, pages 566--577, 2021.

\bibitem[Fan et~al.(2018)Fan, Yang, Hua, Chen, and Wipf]{fan2018revisiting}
Qingnan Fan, Jiaolong Yang, Gang Hua, Baoquan Chen, and David Wipf.
\newblock Revisiting deep intrinsic image decompositions.
\newblock In \emph{Proceedings of the IEEE conference on computer vision and pattern recognition}, pages 8944--8952, 2018.

\bibitem[Gao et~al.(2020)Gao, Chen, Dong, Peers, Xu, and Tong]{gao2020deferred}
Duan Gao, Guojun Chen, Yue Dong, Pieter Peers, Kun Xu, and Xin Tong.
\newblock Deferred neural lighting: free-viewpoint relighting from unstructured photographs.
\newblock \emph{ACM Transactions on Graphics (TOG)}, 39\penalty0 (6):\penalty0 258, 2020.

\bibitem[Garces et~al.(2022)Garces, Rodriguez-Pardo, Casas, and Lopez-Moreno]{garces2022survey}
Elena Garces, Carlos Rodriguez-Pardo, Dan Casas, and Jorge Lopez-Moreno.
\newblock A survey on intrinsic images: Delving deep into lambert and beyond.
\newblock \emph{International Journal of Computer Vision}, 130\penalty0 (3):\penalty0 836--868, 2022.

\bibitem[Gropp et~al.(2020)Gropp, Yariv, Haim, Atzmon, and Lipman]{gropp2020implicit}
Amos Gropp, Lior Yariv, Niv Haim, Matan Atzmon, and Yaron Lipman.
\newblock Implicit geometric regularization for learning shapes.
\newblock In \emph{Proceedings of the 37th International Conference on Machine Learning}, pages 3789--3799, 2020.

\bibitem[Helou et~al.(2020)Helou, Zhou, S{\"u}sstrunk, Timofte, Afifi, Brown, Xu, Cai, Liu, Wang, et~al.]{helou2020aim}
Majed~El Helou, Ruofan Zhou, Sabine S{\"u}sstrunk, Radu Timofte, Mahmoud Afifi, Michael~S Brown, Kele Xu, Hengxing Cai, Yuzhong Liu, Li-Wen Wang, et~al.
\newblock Aim 2020: Scene relighting and illumination estimation challenge.
\newblock \emph{arXiv preprint arXiv:2009.12798}, 2020.

\bibitem[Hou et~al.(2022)Hou, Sarkis, Bi, Tong, and Liu]{hou2022face}
Andrew Hou, Michel Sarkis, Ning Bi, Yiying Tong, and Xiaoming Liu.
\newblock Face relighting with geometrically consistent shadows.
\newblock In \emph{Proceedings of the IEEE/CVF Conference on Computer Vision and Pattern Recognition}, pages 4217--4226, 2022.

\bibitem[Jin et~al.(2023)Jin, Liu, Xu, Zhang, Han, Bi, Zhou, Xu, and Su]{Jin2023TensoIR}
Haian Jin, Isabella Liu, Peijia Xu, Xiaoshuai Zhang, Songfang Han, Sai Bi, Xiaowei Zhou, Zexiang Xu, and Hao Su.
\newblock Tensoir: Tensorial inverse rendering.
\newblock In \emph{Proceedings of the IEEE/CVF Conference on Computer Vision and Pattern Recognition (CVPR)}, 2023.

\bibitem[Kocsis et~al.(2024)Kocsis, Philip, Sunkavalli, Nie{\ss}ner, and Hold-Geoffroy]{kocsis2024lightit}
Peter Kocsis, Julien Philip, Kalyan Sunkavalli, Matthias Nie{\ss}ner, and Yannick Hold-Geoffroy.
\newblock Lightit: Illumination modeling and control for diffusion models.
\newblock In \emph{CVPR}, 2024.

\bibitem[Lettry et~al.(2018)Lettry, Vanhoey, and Van~Gool]{lettry2018unsupervised}
Louis Lettry, Kenneth Vanhoey, and Luc Van~Gool.
\newblock Unsupervised deep single-image intrinsic decomposition using illumination-varying image sequences.
\newblock In \emph{Computer Graphics Forum}, pages 409--419. Wiley Online Library, 2018.

\bibitem[Li and Snavely(2018{\natexlab{a}})]{BigTimeLi18}
Zhengqi Li and Noah Snavely.
\newblock Learning intrinsic image decomposition from watching the world.
\newblock In \emph{Computer Vision and Pattern Recognition (CVPR)}, 2018{\natexlab{a}}.

\bibitem[Li and Snavely(2018{\natexlab{b}})]{li2018cgintrinsics}
Zhengqi Li and Noah Snavely.
\newblock Cgintrinsics: Better intrinsic image decomposition through physically-based rendering.
\newblock In \emph{European Conference on Computer Vision (ECCV)}, 2018{\natexlab{b}}.

\bibitem[Li and Snavely(2018{\natexlab{c}})]{li2018learning}
Zhengqi Li and Noah Snavely.
\newblock Learning intrinsic image decomposition from watching the world.
\newblock In \emph{Proceedings of the IEEE conference on computer vision and pattern recognition}, pages 9039--9048, 2018{\natexlab{c}}.

\bibitem[Li et~al.(2023)Li, M\"uller, Evans, Taylor, Unberath, Liu, and Lin]{li2023neuralangelo}
Zhaoshuo Li, Thomas M\"uller, Alex Evans, Russell~H Taylor, Mathias Unberath, Ming-Yu Liu, and Chen-Hsuan Lin.
\newblock Neuralangelo: High-fidelity neural surface reconstruction.
\newblock In \emph{IEEE Conference on Computer Vision and Pattern Recognition ({CVPR})}, 2023.

\bibitem[Ling et~al.(2022)Ling, Wang, and Xu]{ling2022shadowneus}
Jingwang Ling, Zhibo Wang, and Feng Xu.
\newblock Shadowneus: Neural sdf reconstruction by shadow ray supervision, 2022.

\bibitem[Liu et~al.(2023)Liu, Chen, Fu, Wu, Jin, Li, Wong, Xu, Ramamoorthi, Xu, and Su]{OpenIllumination}
Isabella Liu, Linghao Chen, Ziyang Fu, Liwen Wu, Haian Jin, Zhong Li, Chin Ming~Ryan Wong, Yi Xu, Ravi Ramamoorthi, Zexiang Xu, and Hao Su.
\newblock Openillumination: A multi-illumination dataset for inverse rendering evaluation on real objects.
\newblock In \emph{Advances in Neural Information Processing Systems}, pages 36951--36962. Curran Associates, Inc., 2023.

\bibitem[Liu et~al.(2020)Liu, Li, You, and Lu]{liu2020unsupervised}
Yunfei Liu, Yu Li, Shaodi You, and Feng Lu.
\newblock Unsupervised learning for intrinsic image decomposition from a single image.
\newblock In \emph{Proceedings of the IEEE/CVF Conference on Computer Vision and Pattern Recognition}, pages 3248--3257, 2020.

\bibitem[Loshchilov and Hutter(2018)]{loshchilov2018decoupled}
Ilya Loshchilov and Frank Hutter.
\newblock Decoupled weight decay regularization.
\newblock In \emph{International Conference on Learning Representations}, 2018.

\bibitem[Mildenhall et~al.(2021)Mildenhall, Srinivasan, Tancik, Barron, Ramamoorthi, and Ng]{mildenhall2021nerf}
Ben Mildenhall, Pratul~P Srinivasan, Matthew Tancik, Jonathan~T Barron, Ravi Ramamoorthi, and Ren Ng.
\newblock Nerf: Representing scenes as neural radiance fields for view synthesis.
\newblock \emph{Communications of the ACM}, 65\penalty0 (1):\penalty0 99--106, 2021.

\bibitem[M\"uller et~al.(2022)M\"uller, Evans, Schied, and Keller]{mueller2022instant}
Thomas M\"uller, Alex Evans, Christoph Schied, and Alexander Keller.
\newblock Instant neural graphics primitives with a multiresolution hash encoding.
\newblock \emph{ACM Trans. Graph.}, 41\penalty0 (4):\penalty0 102:1--102:15, 2022.

\bibitem[Murmann et~al.(2019)Murmann, Gharbi, Aittala, and Durand]{murmann2019dataset}
Lukas Murmann, Michael Gharbi, Miika Aittala, and Fredo Durand.
\newblock A dataset of multi-illumination images in the wild.
\newblock In \emph{Proceedings of the IEEE/CVF International Conference on Computer Vision}, pages 4080--4089, 2019.

\bibitem[Nestmeyer et~al.(2020)Nestmeyer, Lalonde, Matthews, and Lehrmann]{nestmeyer2020learning}
Thomas Nestmeyer, Jean-Fran{\c{c}}ois Lalonde, Iain Matthews, and Andreas Lehrmann.
\newblock Learning physics-guided face relighting under directional light.
\newblock In \emph{Proceedings of the IEEE/CVF Conference on Computer Vision and Pattern Recognition}, pages 5124--5133, 2020.

\bibitem[Pandey et~al.(2021)Pandey, Orts-Escolano, Legendre, Haene, Bouaziz, Rhemann, Debevec, and Fanello]{pandey2021total}
Rohit Pandey, Sergio Orts-Escolano, Chloe Legendre, Christian Haene, Sofien Bouaziz, Christoph Rhemann, Paul~E Debevec, and Sean~Ryan Fanello.
\newblock Total relighting: learning to relight portraits for background replacement.
\newblock \emph{ACM Trans. Graph.}, 40:\penalty0 43--1, 2021.

\bibitem[Pedregosa et~al.(2011)Pedregosa, Varoquaux, Gramfort, Michel, Thirion, Grisel, Blondel, Prettenhofer, Weiss, Dubourg, Vanderplas, Passos, Cournapeau, Brucher, Perrot, and Duchesnay]{scikit-learn}
F. Pedregosa, G. Varoquaux, A. Gramfort, V. Michel, B. Thirion, O. Grisel, M. Blondel, P. Prettenhofer, R. Weiss, V. Dubourg, J. Vanderplas, A. Passos, D. Cournapeau, M. Brucher, M. Perrot, and E. Duchesnay.
\newblock Scikit-learn: Machine learning in {P}ython.
\newblock \emph{Journal of Machine Learning Research}, 12:\penalty0 2825--2830, 2011.

\bibitem[Puthussery et~al.(2020)Puthussery, Kuriakose, C~V, et~al.]{puthussery2020wdrn}
Densen Puthussery, Melvin Kuriakose, Jiji C~V, et~al.
\newblock Wdrn: A wavelet decomposed relightnet for image relighting.
\newblock \emph{arXiv preprint arXiv:2009.06678}, 2020.

\bibitem[Rudnev et~al.(2022)Rudnev, Elgharib, Smith, Liu, Golyanik, and Theobalt]{rudnev2022nerfosr}
Viktor Rudnev, Mohamed Elgharib, William Smith, Lingjie Liu, Vladislav Golyanik, and Christian Theobalt.
\newblock Nerf for outdoor scene relighting.
\newblock In \emph{European Conference on Computer Vision (ECCV)}, 2022.

\bibitem[Srinivasan et~al.(2021)Srinivasan, Deng, Zhang, Tancik, Mildenhall, and Barron]{srinivasan2021nerv}
Pratul~P Srinivasan, Boyang Deng, Xiuming Zhang, Matthew Tancik, Ben Mildenhall, and Jonathan~T Barron.
\newblock Nerv: Neural reflectance and visibility fields for relighting and view synthesis.
\newblock In \emph{Proceedings of the IEEE/CVF Conference on Computer Vision and Pattern Recognition}, pages 7495--7504, 2021.

\bibitem[Sun et~al.(2019)Sun, Barron, Tsai, Xu, Yu, Fyffe, Rhemann, Busch, Debevec, and Ramamoorthi]{sun2019single}
Tiancheng Sun, Jonathan~T Barron, Yun-Ta Tsai, Zexiang Xu, Xueming Yu, Graham Fyffe, Christoph Rhemann, Jay Busch, Paul~E Debevec, and Ravi Ramamoorthi.
\newblock Single image portrait relighting.
\newblock \emph{ACM Trans. Graph.}, 38\penalty0 (4):\penalty0 79--1, 2019.

\bibitem[Toschi et~al.(2023)Toschi, De~Matteo, Spezialetti, De~Gregorio, Di~Stefano, and Salti]{Toschi_2023_CVPR}
Marco Toschi, Riccardo De~Matteo, Riccardo Spezialetti, Daniele De~Gregorio, Luigi Di~Stefano, and Samuele Salti.
\newblock Relight my nerf: A dataset for novel view synthesis and relighting of real world objects.
\newblock In \emph{Proceedings of the IEEE/CVF Conference on Computer Vision and Pattern Recognition (CVPR)}, pages 20762--20772, 2023.

\bibitem[Wang et~al.(2020)Wang, Siu, Liu, Li, and Lun]{wang2020deep}
Li-Wen Wang, Wan-Chi Siu, Zhi-Song Liu, Chu-Tak Li, and Daniel~PK Lun.
\newblock Deep relighting networks for image light source manipulation.
\newblock \emph{arXiv preprint arXiv:2008.08298}, 2020.

\bibitem[Wang et~al.(2023)Wang, Wu, and Xu]{wang2023udcnerf}
Yuxin Wang, Wayne Wu, and Dan Xu.
\newblock Learning unified decompositional and compositional nerf for editable novel view synthesis.
\newblock In \emph{ICCV}, 2023.

\bibitem[Wang et~al.(2004)Wang, Bovik, Sheikh, and Simoncelli]{wang2004image}
Zhou Wang, Alan~C Bovik, Hamid~R Sheikh, and Eero~P Simoncelli.
\newblock Image quality assessment: from error visibility to structural similarity.
\newblock \emph{IEEE transactions on image processing}, 13\penalty0 (4):\penalty0 600--612, 2004.

\bibitem[Yang et~al.(2023)Yang, Chen, Gao, Yuan, Wu, Zhou, and Jin]{yang2023sireir}
Ziyi Yang, Yanzhen Chen, Xinyu Gao, Yazhen Yuan, Yu Wu, Xiaowei Zhou, and Xiaogang Jin.
\newblock Sire-ir: Inverse rendering for brdf reconstruction with shadow and illumination removal in high-illuminance scenes.
\newblock \emph{arXiv preprint arXiv:2310.13030}, 2023.

\bibitem[Ye et~al.(2023)Ye, Chen, Bao, Bao, Pollefeys, Cui, and Zhang]{Ye2023IntrinsicNeRF}
Weicai Ye, Shuo Chen, Chong Bao, Hujun Bao, Marc Pollefeys, Zhaopeng Cui, and Guofeng Zhang.
\newblock {IntrinsicNeRF: Learning Intrinsic Neural Radiance Fields for Editable Novel View Synthesis}.
\newblock In \emph{Proceedings of the IEEE/CVF International Conference on Computer Vision}, 2023.

\bibitem[Zeng et~al.(2023)Zeng, Chen, Dong, Peers, Wu, and Tong]{zeng2023nrhints}
Chong Zeng, Guojun Chen, Yue Dong, Pieter Peers, Hongzhi Wu, and Xin Tong.
\newblock Relighting neural radiance fields with shadow and highlight hints.
\newblock In \emph{ACM SIGGRAPH 2023 Conference Proceedings}, 2023.

\bibitem[Zhang et~al.(2021{\natexlab{a}})Zhang, Luan, Wang, Bala, and Snavely]{physg2021}
Kai Zhang, Fujun Luan, Qianqian Wang, Kavita Bala, and Noah Snavely.
\newblock {PhySG}: {I}nverse rendering with spherical gaussians for physics-based material editing and relighting.
\newblock In \emph{The IEEE/CVF Conference on Computer Vision and Pattern Recognition (CVPR)}, 2021{\natexlab{a}}.

\bibitem[Zhang et~al.(2018)Zhang, Isola, Efros, Shechtman, and Wang]{zhang2018unreasonable}
Richard Zhang, Phillip Isola, Alexei~A Efros, Eli Shechtman, and Oliver Wang.
\newblock The unreasonable effectiveness of deep features as a perceptual metric.
\newblock In \emph{Proceedings of the IEEE conference on computer vision and pattern recognition}, pages 586--595, 2018.

\bibitem[Zhang et~al.(2021{\natexlab{b}})Zhang, Srinivasan, Deng, Debevec, Freeman, and Barron]{zhang2021nerfactor}
Xiuming Zhang, Pratul~P Srinivasan, Boyang Deng, Paul Debevec, William~T Freeman, and Jonathan~T Barron.
\newblock Nerfactor: Neural factorization of shape and reflectance under an unknown illumination.
\newblock \emph{ACM Transactions on Graphics (ToG)}, 40\penalty0 (6):\penalty0 1--18, 2021{\natexlab{b}}.

\bibitem[Zhang et~al.(2022)Zhang, Sun, He, Fu, Jia, and Zhou]{zhang2022invrender}
Yuanqing Zhang, Jiaming Sun, Xingyi He, Huan Fu, Rongfei Jia, and Xiaowei Zhou.
\newblock Modeling indirect illumination for inverse rendering.
\newblock In \emph{CVPR}, 2022.

\bibitem[Zhou et~al.(2019)Zhou, Hadap, Sunkavalli, and Jacobs]{zhou2019deep}
Hao Zhou, Sunil Hadap, Kalyan Sunkavalli, and David~W Jacobs.
\newblock Deep single-image portrait relighting.
\newblock In \emph{Proceedings of the IEEE International Conference on Computer Vision}, pages 7194--7202, 2019.

\end{thebibliography}
}

\clearpage
\setcounter{page}{1}
\maketitlesupplementary

\noindent In this supplementary material, we present the following:
\begin{enumerate}
    \item More results on the Synthetic Dataset.
    \item More results on the Real Object Dataset.    
    \item More results on the ReNe Dataset.
\end{enumerate}
We have also submitted a supplementary video to showcase the results of our method on all datasets.

\section{More results on the Synthetic Dataset}
\begin{figure*}[!ht]
\centering
\includegraphics[width=\linewidth]{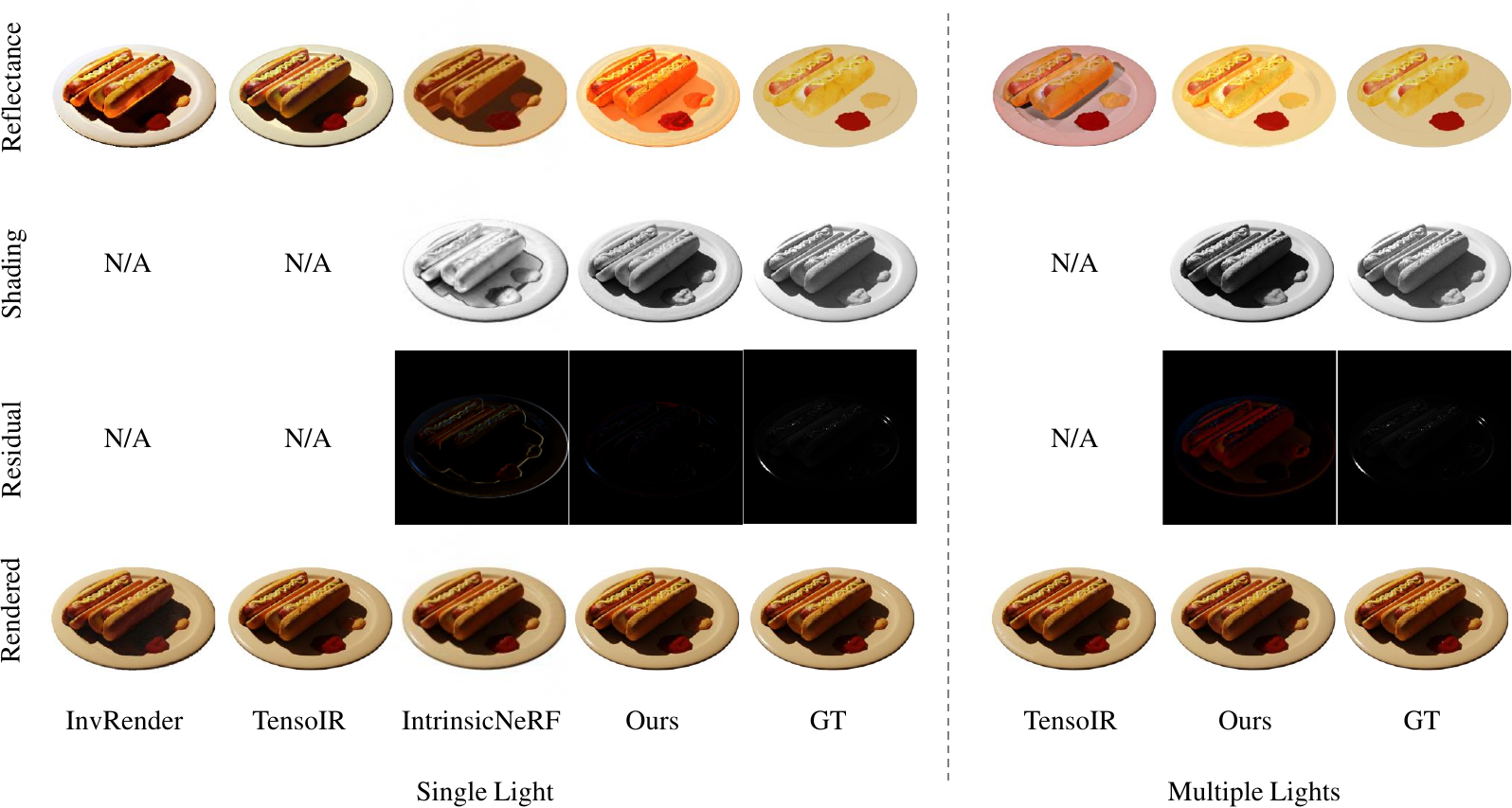}
\vspace{1mm}
\\
\includegraphics[width=0.617\linewidth]{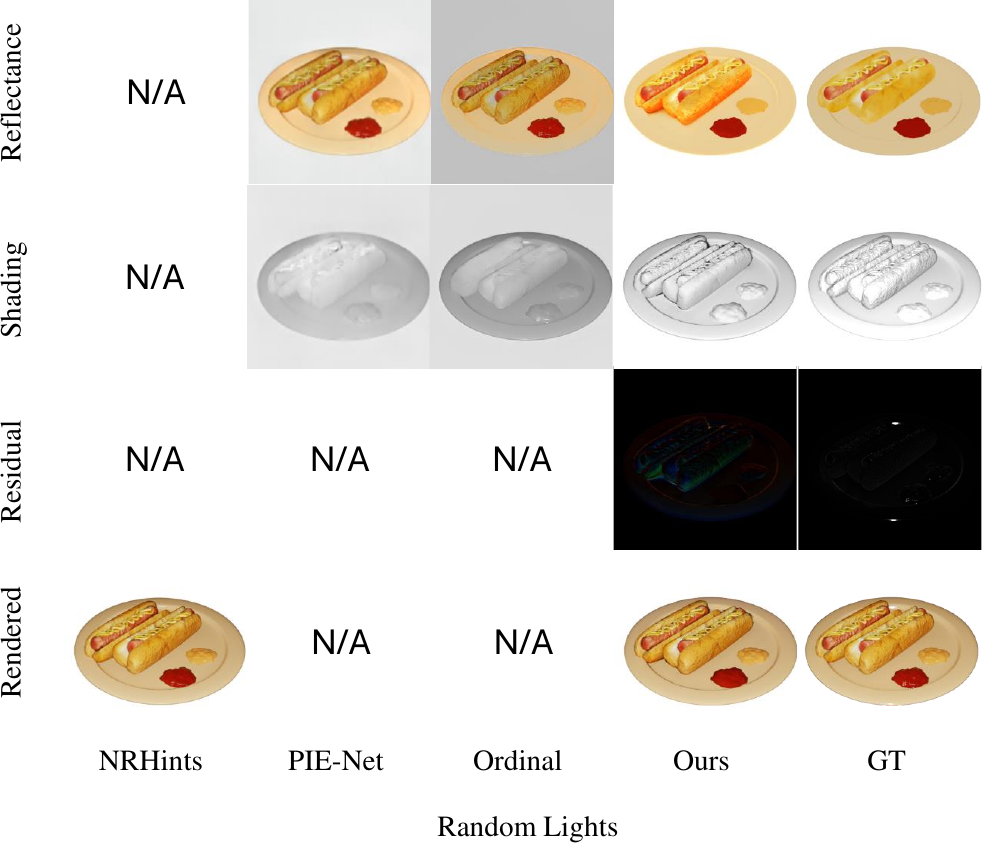}
\caption{Additional Qualitative Results on the Synthetic Dataset. (Hotdog)}
\label{fig: sup_syn_1 Hotdog}
\end{figure*}

\begin{figure*}[!ht]
\centering
\includegraphics[width=\linewidth]{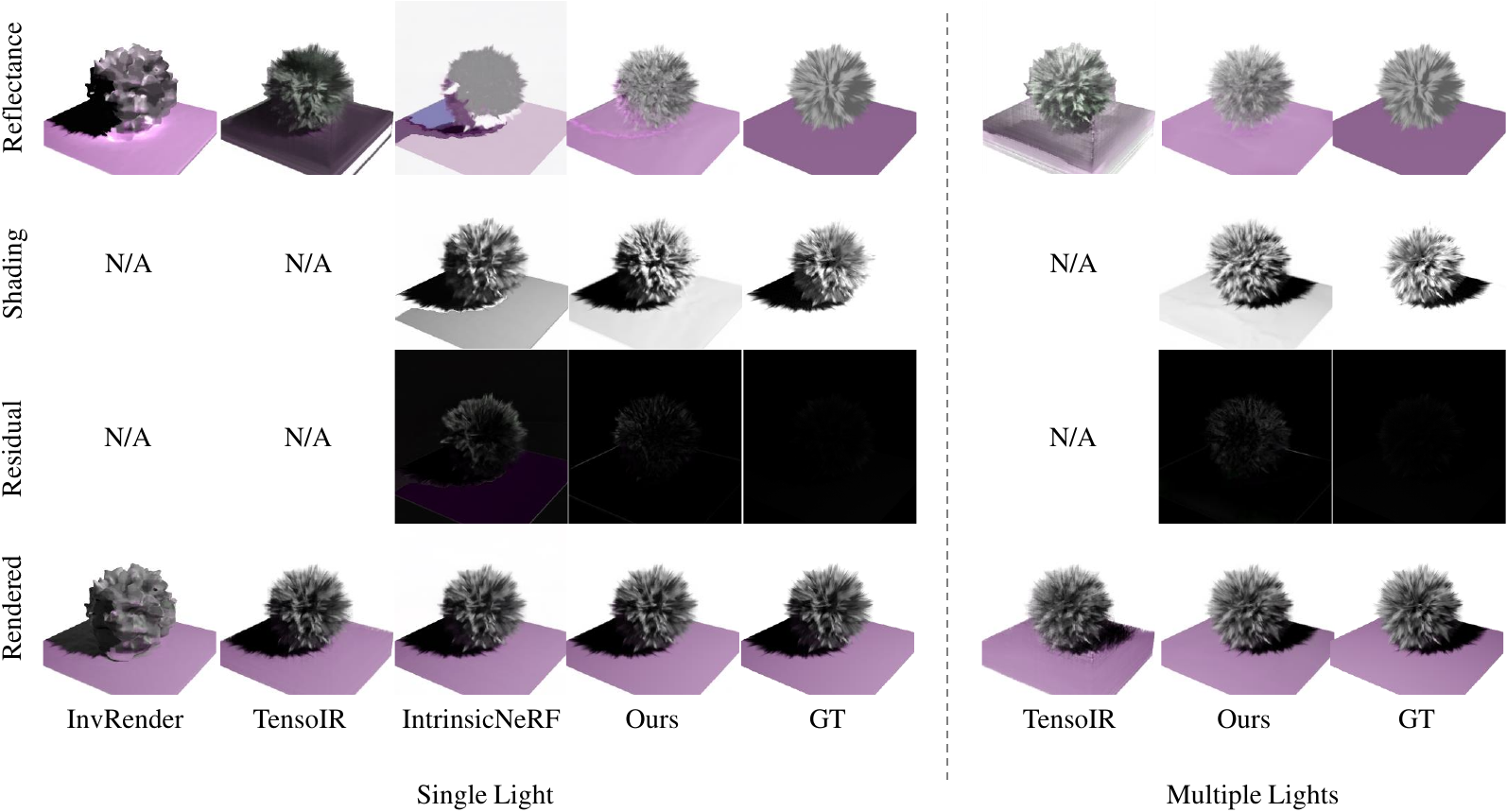}
\vspace{1mm}
\\
\includegraphics[width=0.617\linewidth]{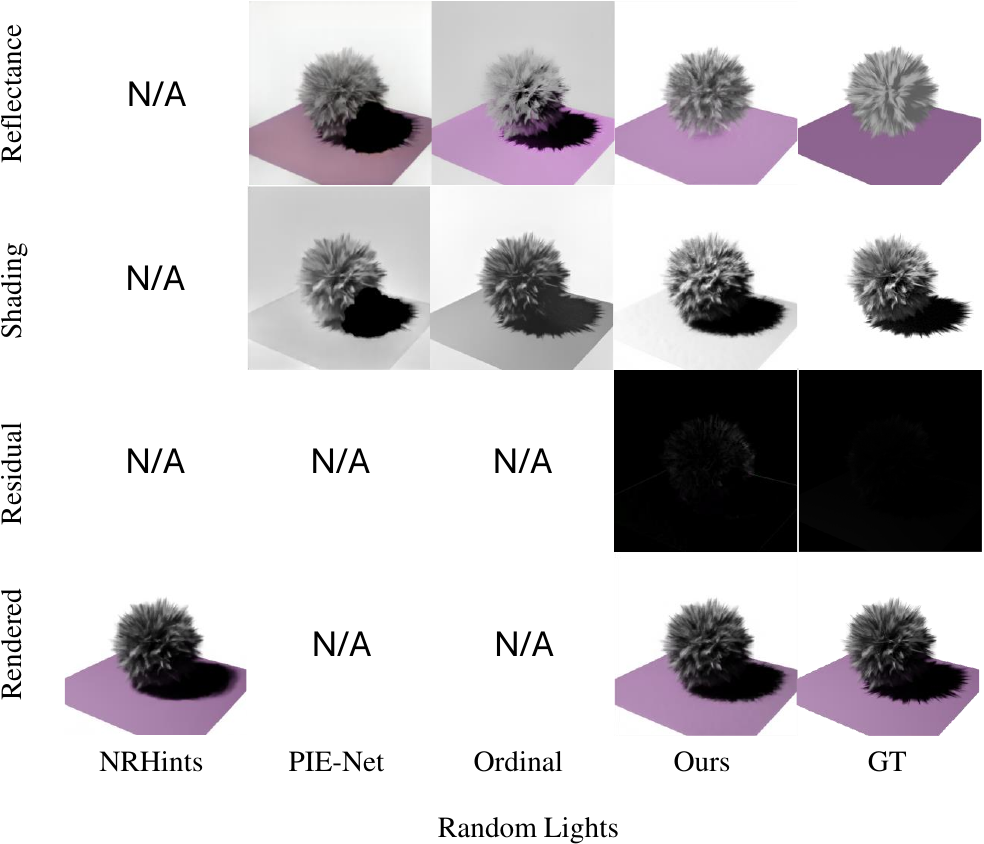}
\caption{Additional Qualitative Results on the Synthetic Dataset (FurBall). However, it is worth noting that in the GT, the shading of the ground in this model was not correctly rendered in Blender. Our method under the Random Lights setting achieved the most reasonable results.}
\label{fig: sup_syn_1 FurBall}
\end{figure*}

\begin{figure*}[!ht]
\centering
\includegraphics[width=\linewidth]{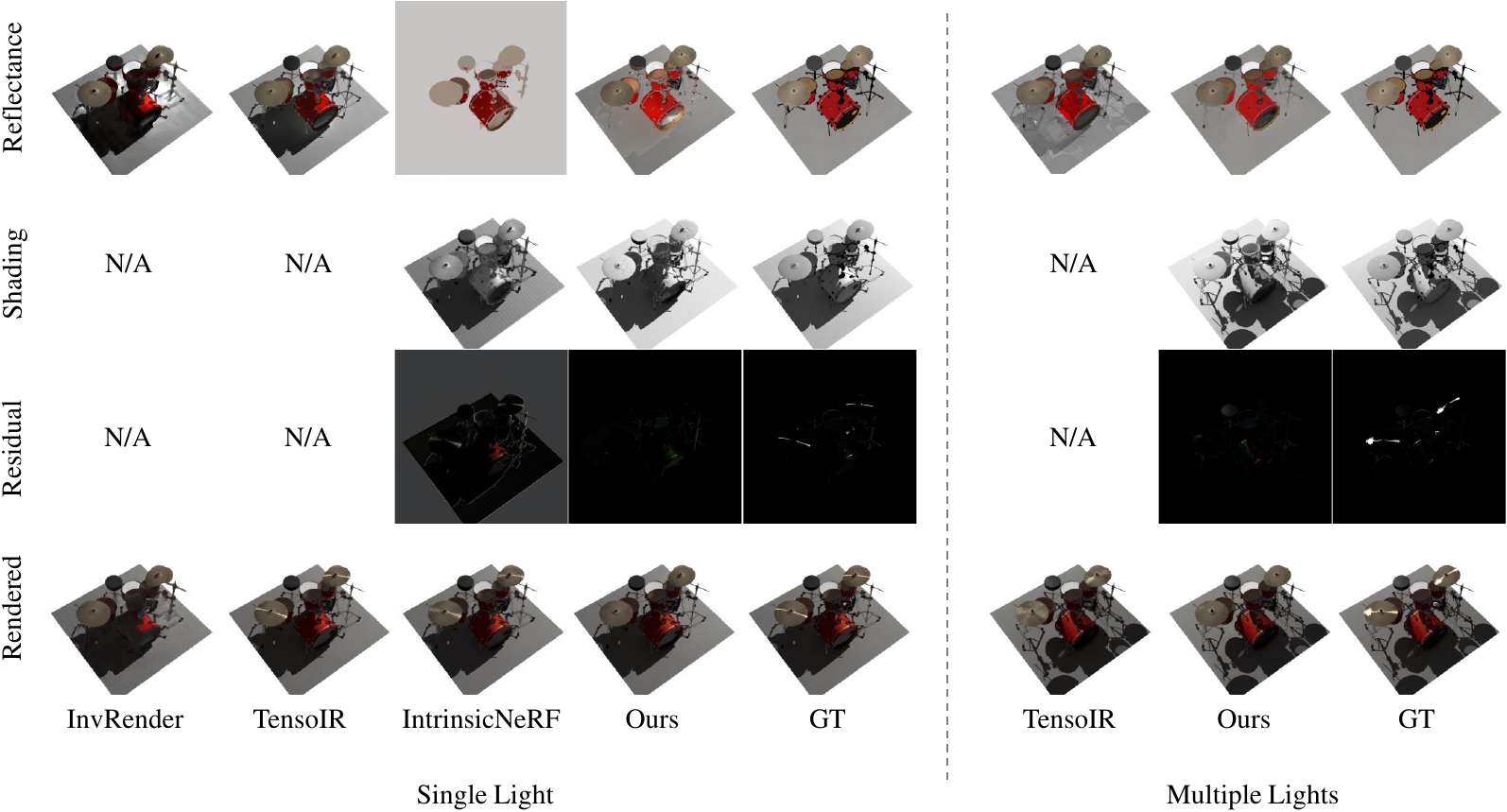}
\vspace{1mm}
\\
\includegraphics[width=0.617\linewidth]{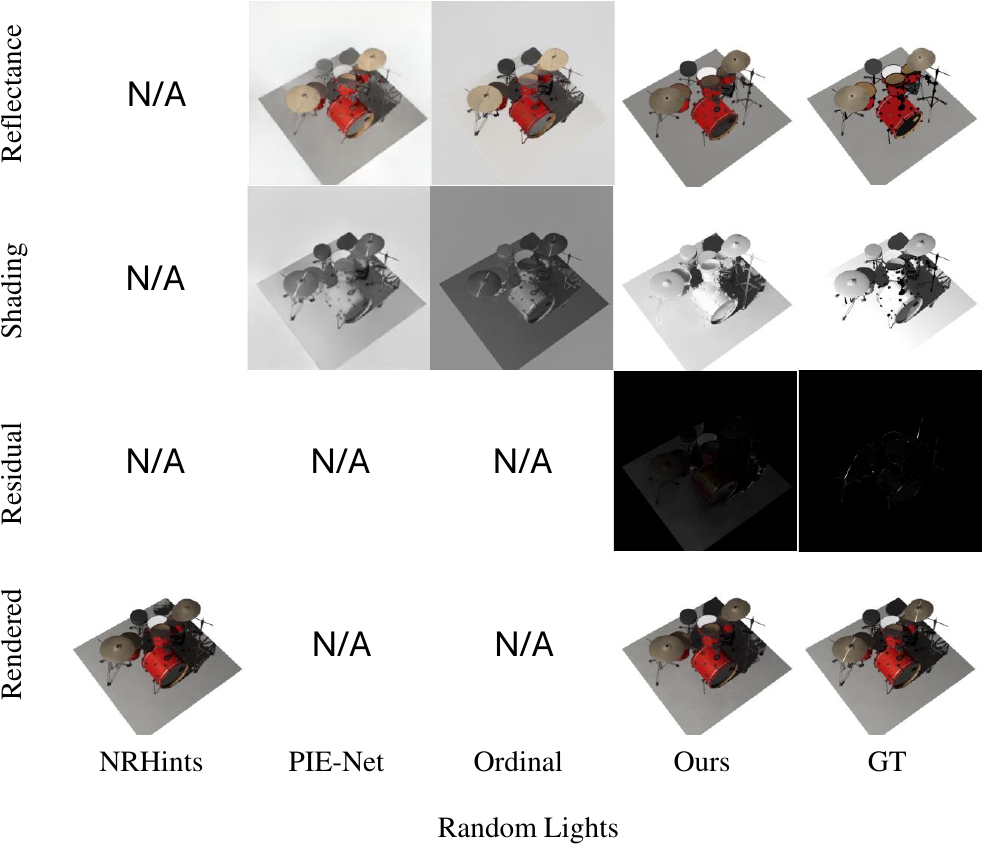}
\caption{Additional Qualitative Results on the Synthetic Dataset. (Drums)}
\label{fig: sup_syn_1 Drums}
\end{figure*}

\begin{figure*}[!ht]
\centering
\includegraphics[width=\linewidth]{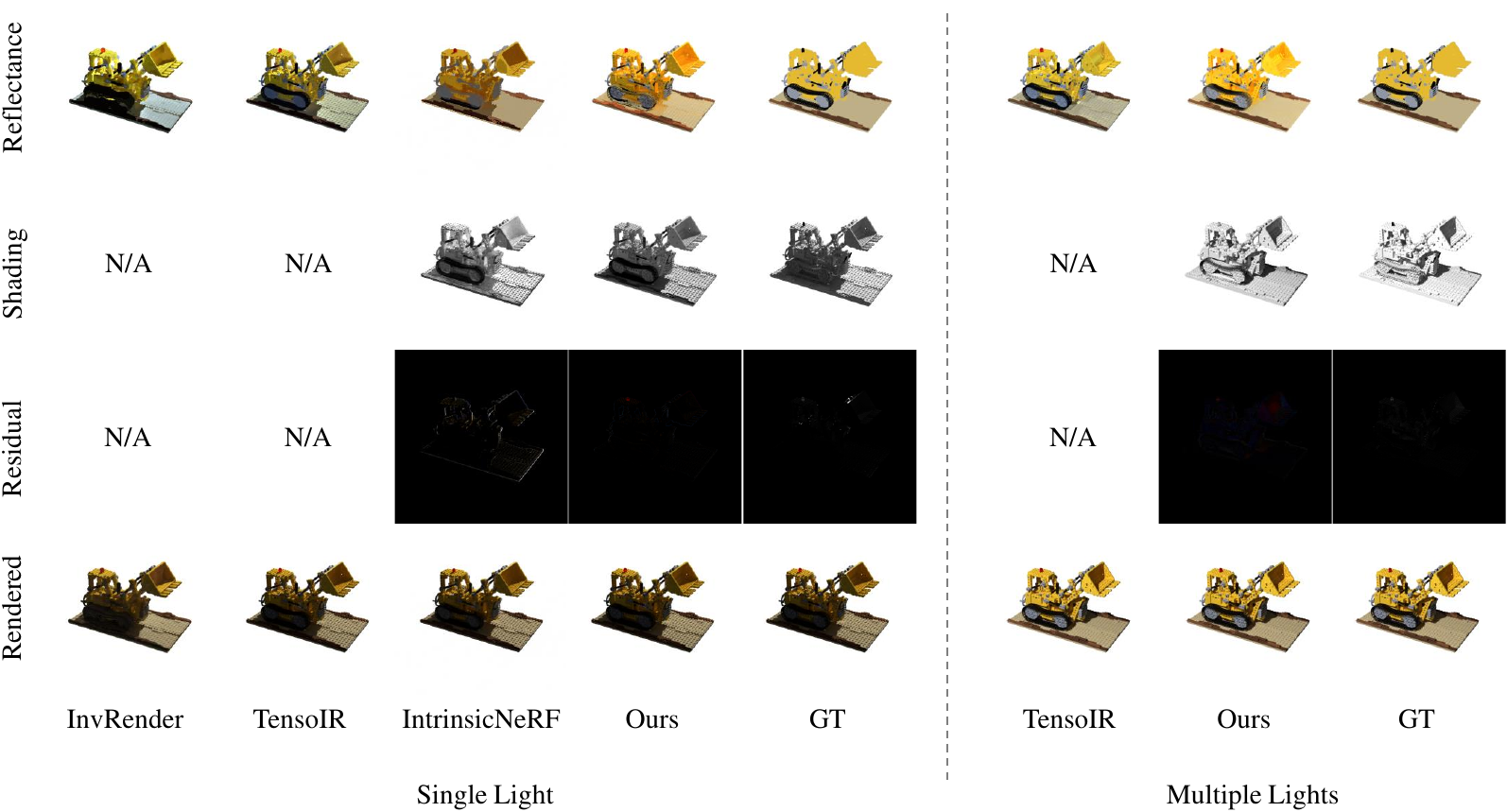}
\vspace{1mm}
\\
\includegraphics[width=0.617\linewidth]{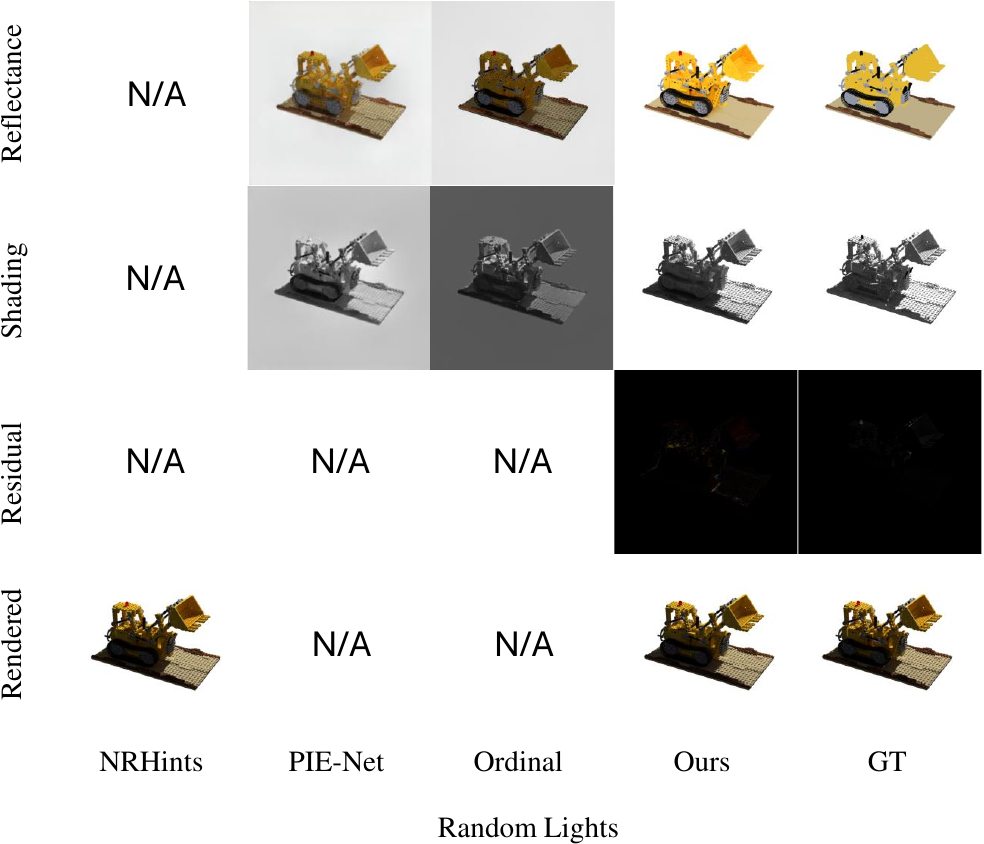}
\caption{Additional Qualitative Results on the Synthetic Dataset. (Lego)}
\label{fig: sup_syn_1 Lego}
\end{figure*}

From \cref{fig: sup_syn_1 Hotdog} to \cref{fig: sup_syn_1 Lego}, we present additional qualitative results with all settings including Single Light, Multiple Lights, and Random Lights on the Synthetic Dataset.

In the Synthetic Dataset, we conduct experiments on four scenes: Hotdog, FurBall, Drums, and Lego. Across all scenes, we observe that under the original Random Lights setting, our method achieves best results, closely matching the GT. The predicted reflectance and shading are significantly better than those of PIE-Net \cite{dasPIENet} and Ordinal \cite{careagaIntrinsic}, and the relighting results are comparable to NRHints \cite{zeng2023nrhints}.

Under the Single Light and Multiple Light settings, we compare our method with additional approaches. In the Single Light setting, all methods struggle to consistently produce good intrinsic images, particularly in separating cast shadows. However, our method successfully removes most cast shadows in the predicted reflectance for scenes like FurBall, Drums, and Lego, outperforming other methods. This underscores the importance of multi-light information for intrinsic decomposition.

In the Multiple Lights setting (with four different light positions), our method achieves satisfactory results, very close to the GT and the Random Lights setting, demonstrating that four different light positions are sufficient for our method to perform well. Under the same setting, TensoIR \cite{Jin2023TensoIR} fails to produce satisfactory results, with residual cast shadows mixing into the reflectance.

Overall, since the camera view used in comparisons is the same, the reflectance should have the same GT across all settings. Our predicted reflectance consistently outperforms others in comparisons within the same rows, while also showing improvements as the number of light positions increases.

\section{More results on the Real Object Dataset}
\begin{figure*}[!ht]
\centering
\includegraphics[width=0.8\linewidth]{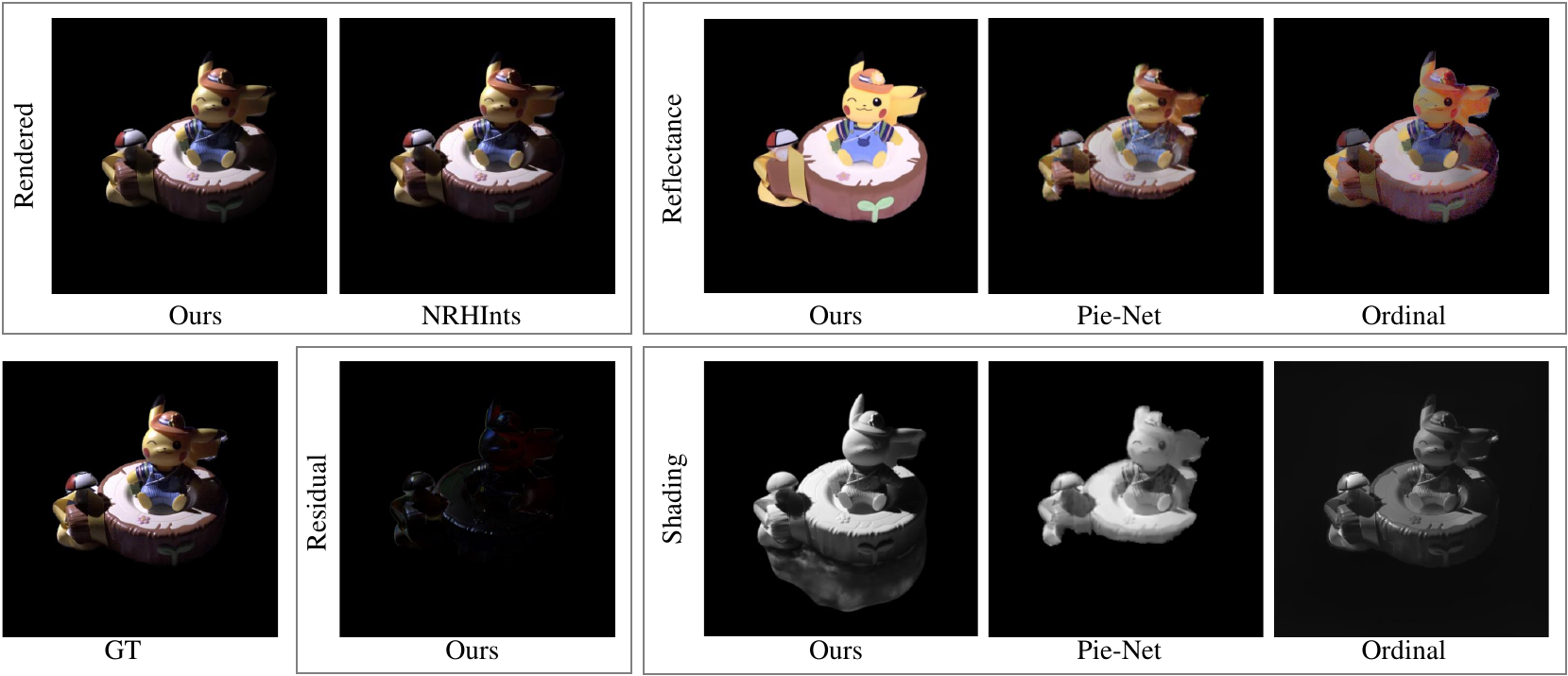} \\
\includegraphics[width=0.8\linewidth]{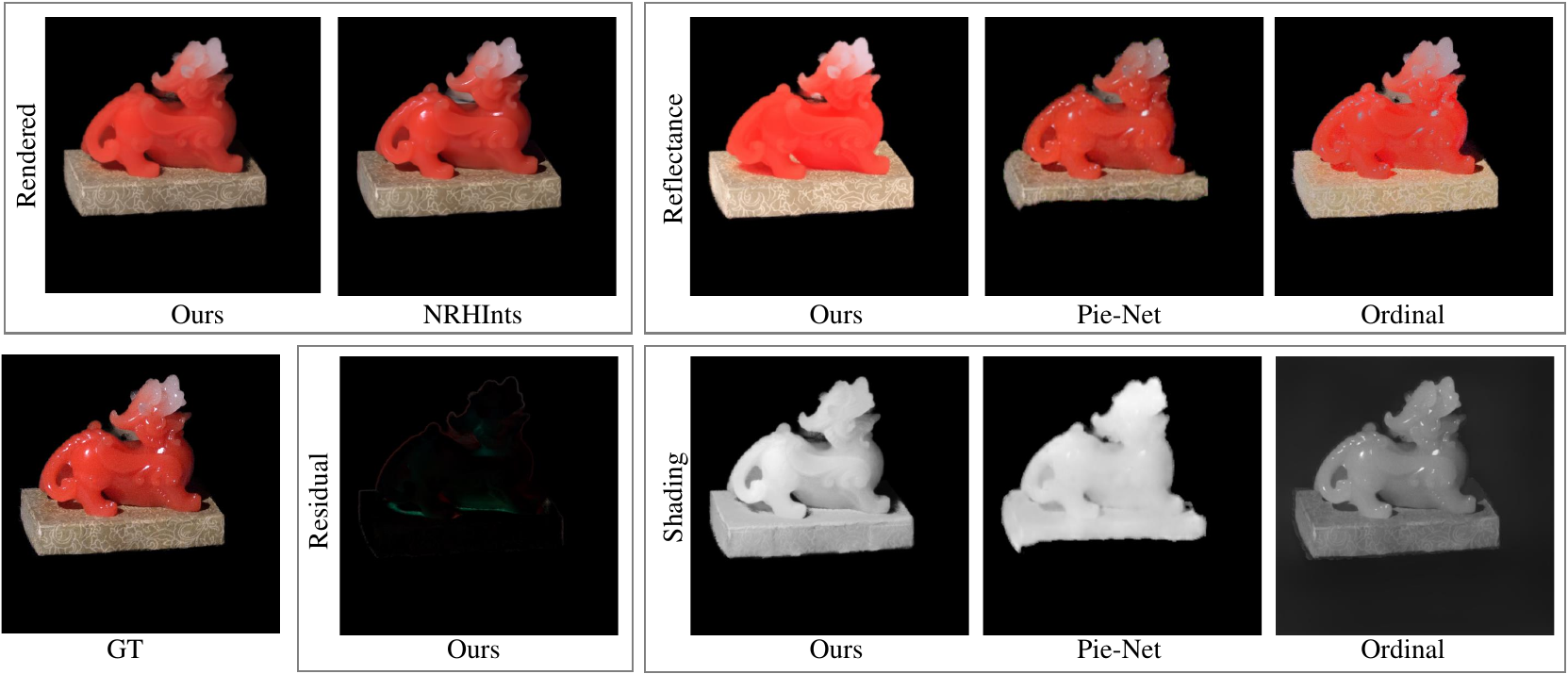}
\includegraphics[width=0.8\linewidth]{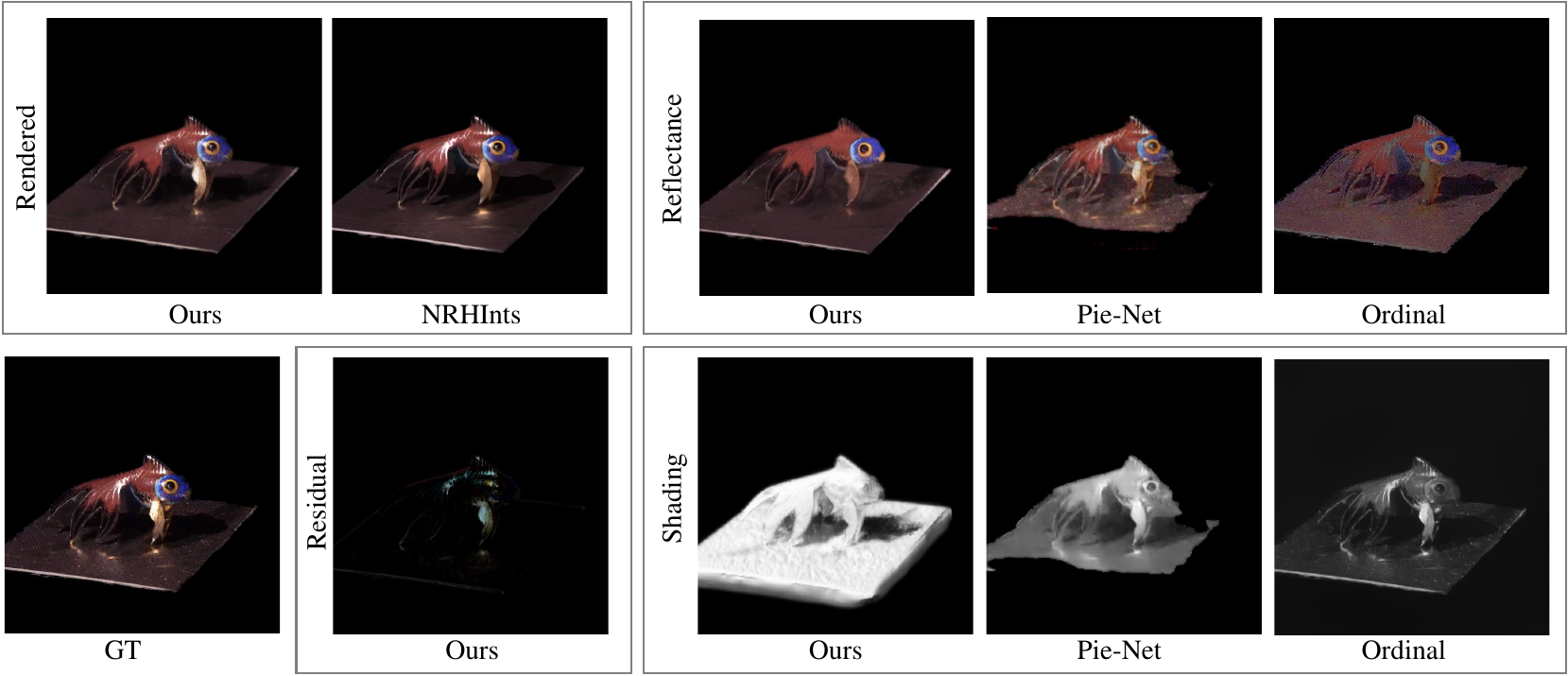}
\caption{Additional Qualitative Results on the Real Object Dataset.}
\label{fig: sup_real_comparison}
\end{figure*}

\begin{figure*}[!ht]
\centering
\includegraphics[width=\linewidth]{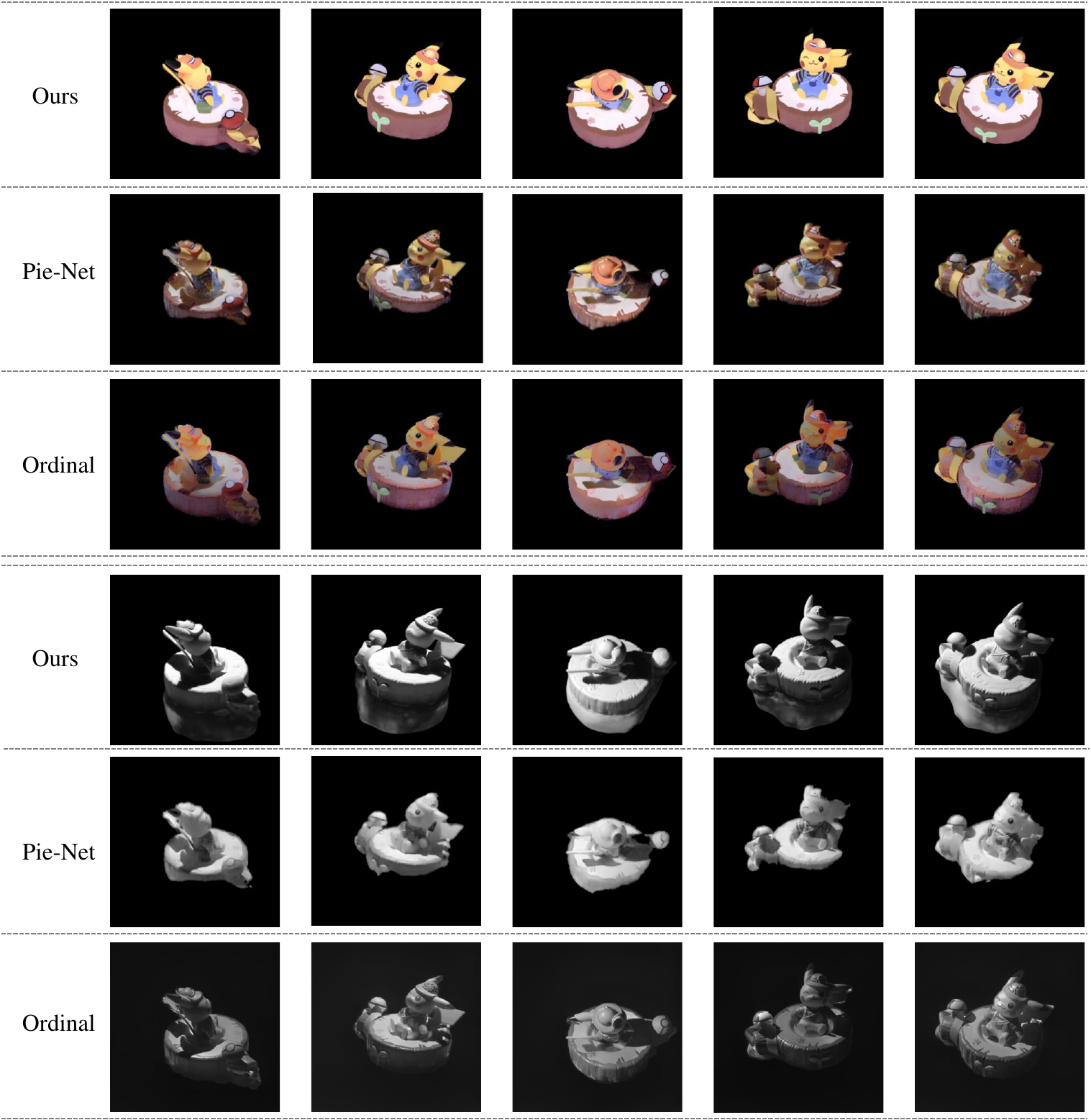}
\caption{Reflectance and Shading estimation by our method for different points of view of the same scene on the Real Object Dataset.}
\label{fig: sup_real_pview}
\end{figure*}

We present different results for the 4 scenes of the Real Object Dataset, including Fish, Pikachu, Pixiu, and FurScene \cite{gao2020deferred, zeng2023nrhints}. In \cref{fig: sup_real_comparison}, we show additional results on further scenes. In \cref{fig: sup_real_pview}, we present the results of reflectance and shading from multiple camera views and light positions to demonstrate the coherence of the approach.

\begin{figure*}[!ht]
\centering
\includegraphics[width=0.8\linewidth]{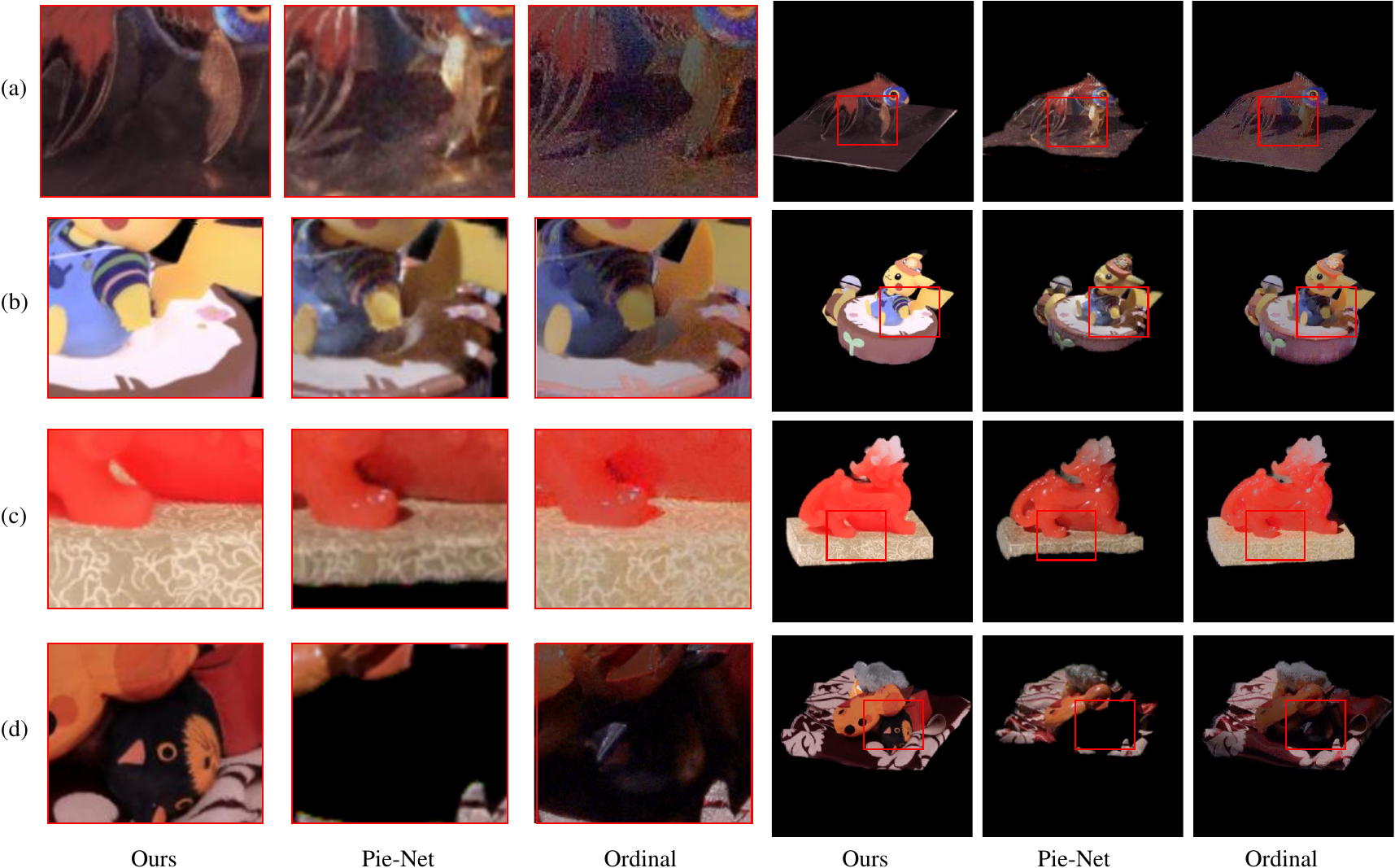}
\caption{Reflectance Estimation details for different Scenes of the Real Object Dataset. }
\label{fig: sup_real_refzoom}
\end{figure*}

\begin{figure*}[!ht]
\centering
\includegraphics[width=0.8\linewidth]{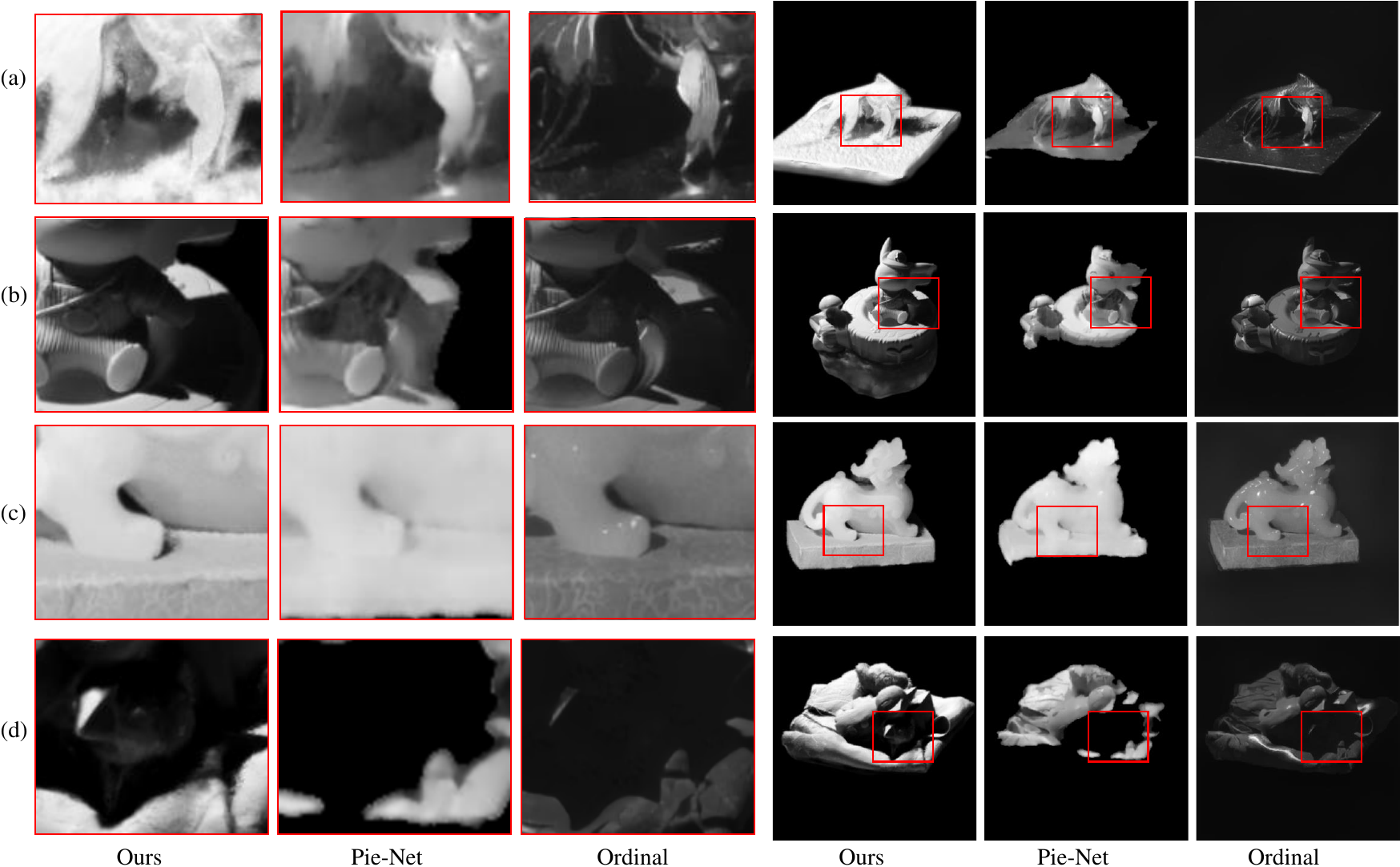}
\caption{Shading Estimation details for different Scenes of the Real Object Dataset. }
\label{fig: sup_real_shazoom}
\end{figure*}

In \cref{fig: sup_real_refzoom} and \cref{fig: sup_real_shazoom}, we zoom in on certain areas to show more details of the results. For reflectance, we want to highlight rows (c) and (d), where our reflectance shows very promising results. In (c), our approach correctly estimates the reflectance, likely due to the robust 3D information considered in the pseudo-shading. In (d), the high quality of the result is probably due to the pseudo-reflectance labels that effectively represent the reflectance in low-light areas. However, a limitation of our approach is evident in the shading of row (a). Although the global surface shading is satisfactory, the edges are not sharp, and there is some confusion in the areas below the object, likely due to none of the viewpoints or light sources providing adequate information for proper reconstruction.

For completeness of the comparison and potential needs, we also compared the relighting results with NRHints \cite{zeng2023nrhints}. A quantitative evaluation is provided in \cref{tab: real relight}. As shown, metrics show our method presents slightly lower PSNR and SSIM but better LPIPS.  

\begin{table}[!ht]
\centering
\begin{tabular}{ccccc}
\hline
        & PSNR$\uparrow$  & SSIM$\uparrow$   & LPIPS$\downarrow$                & MSE $\downarrow$  \\
        \hline
NRHints & \textbf{31.62} & \textbf{0.9623} & 0.0997 &  ---   \\
Ours    & 30.55 & 0.9341 & \textbf{0.0797} & 0.0012  \\
\hline
\end{tabular}
\caption{Quantitative results of the novel view synthesis and relighting on the \textbf{Real Object Dataset}.}
\label{tab: real relight}
\vspace{-3mm}
\end{table}

\section{More results on the ReNe Dataset}

\begin{figure*}[!ht]
\centering
\includegraphics[width=1.0\linewidth]{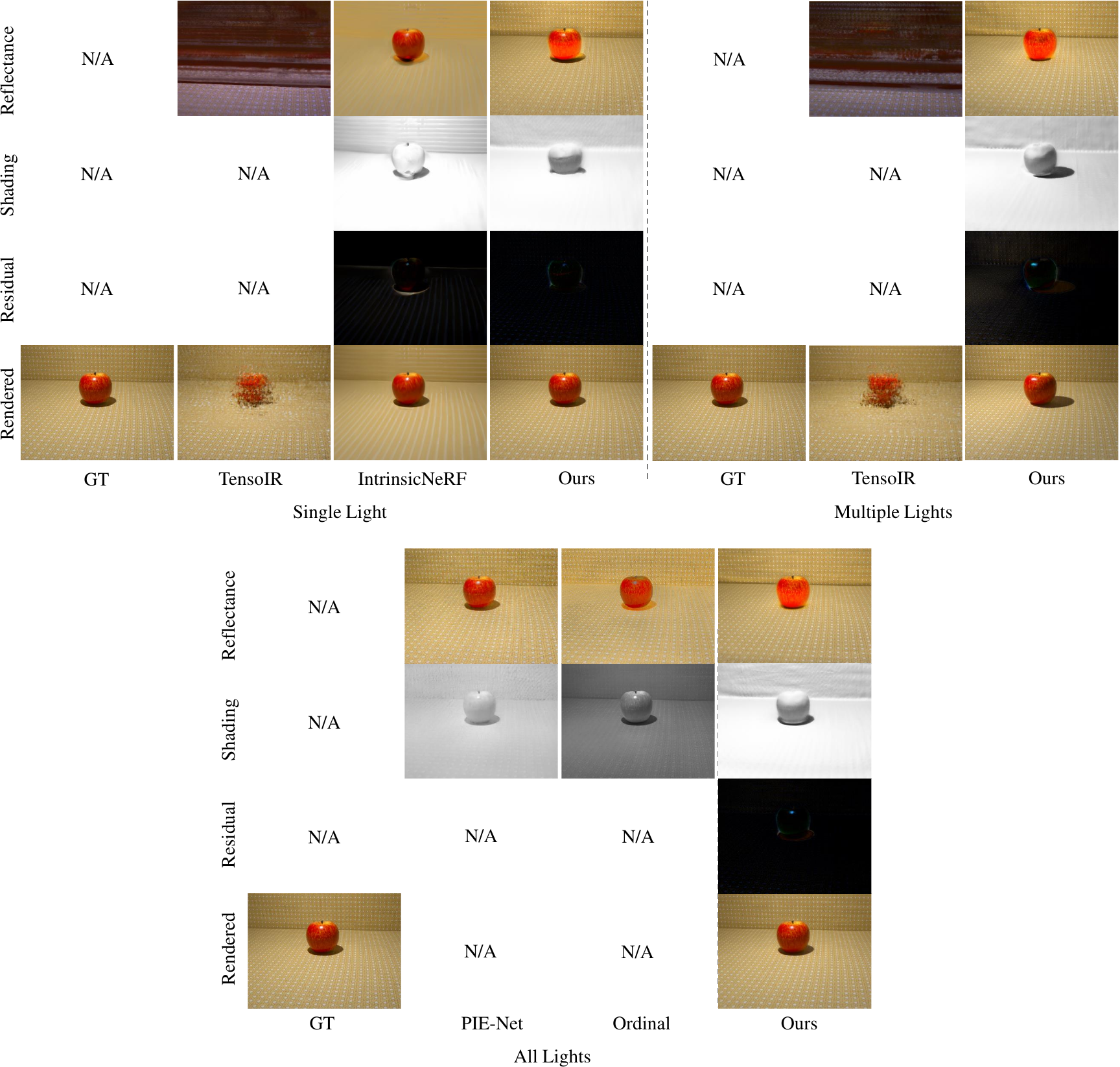}
\caption{Additional Qualitative Results on the ReNe dataset. (Apple).}
\label{suprene: Apple}
\end{figure*}

\begin{figure*}[!ht]
\centering
\includegraphics[width=1.0\linewidth]{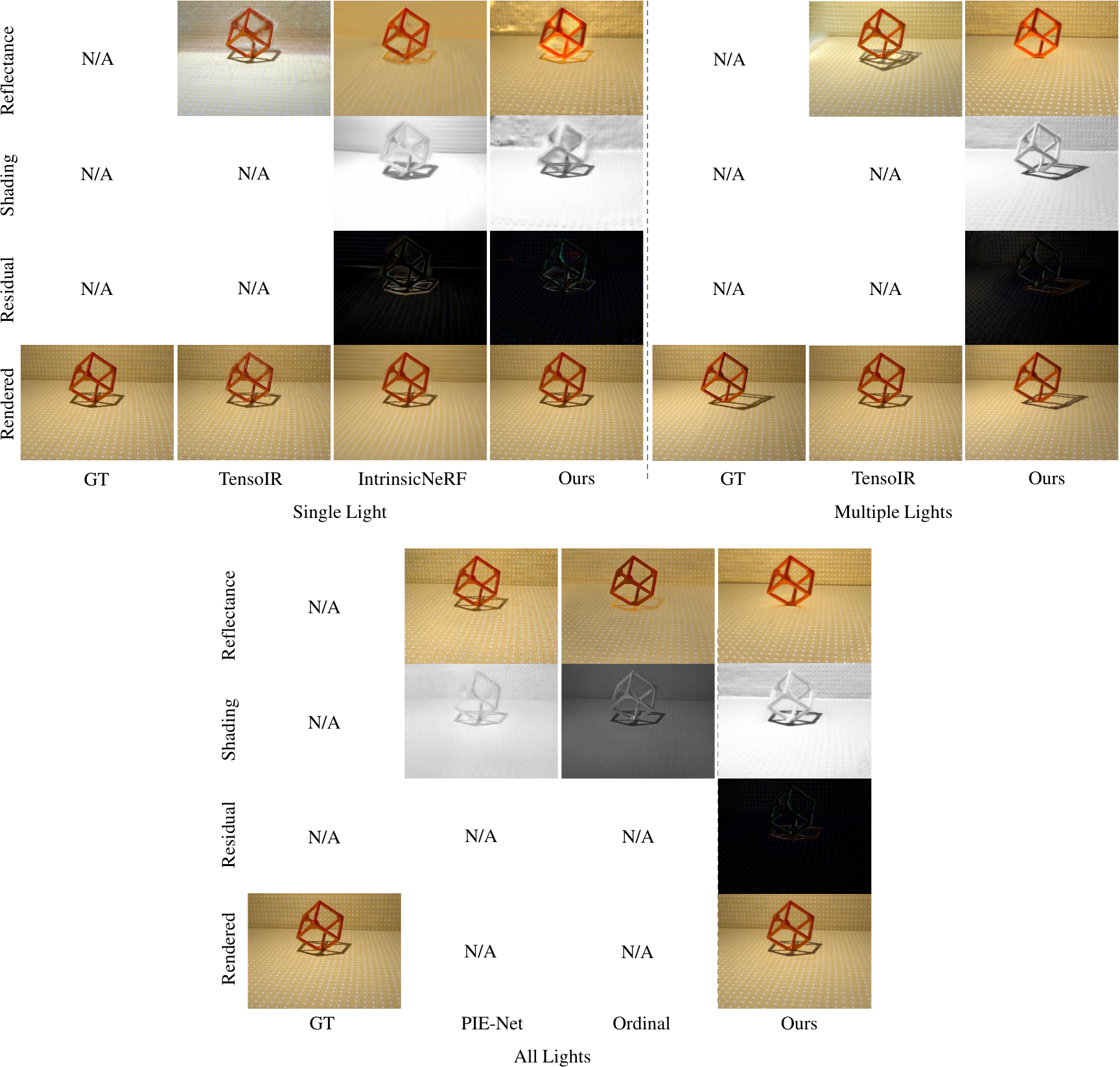}
\caption{Additional Qualitative Results on the ReNe dataset. (Cube).}
\label{suprene: Cube}
\end{figure*}

\begin{figure*}[!ht]
\centering
\includegraphics[width=1.0\linewidth]{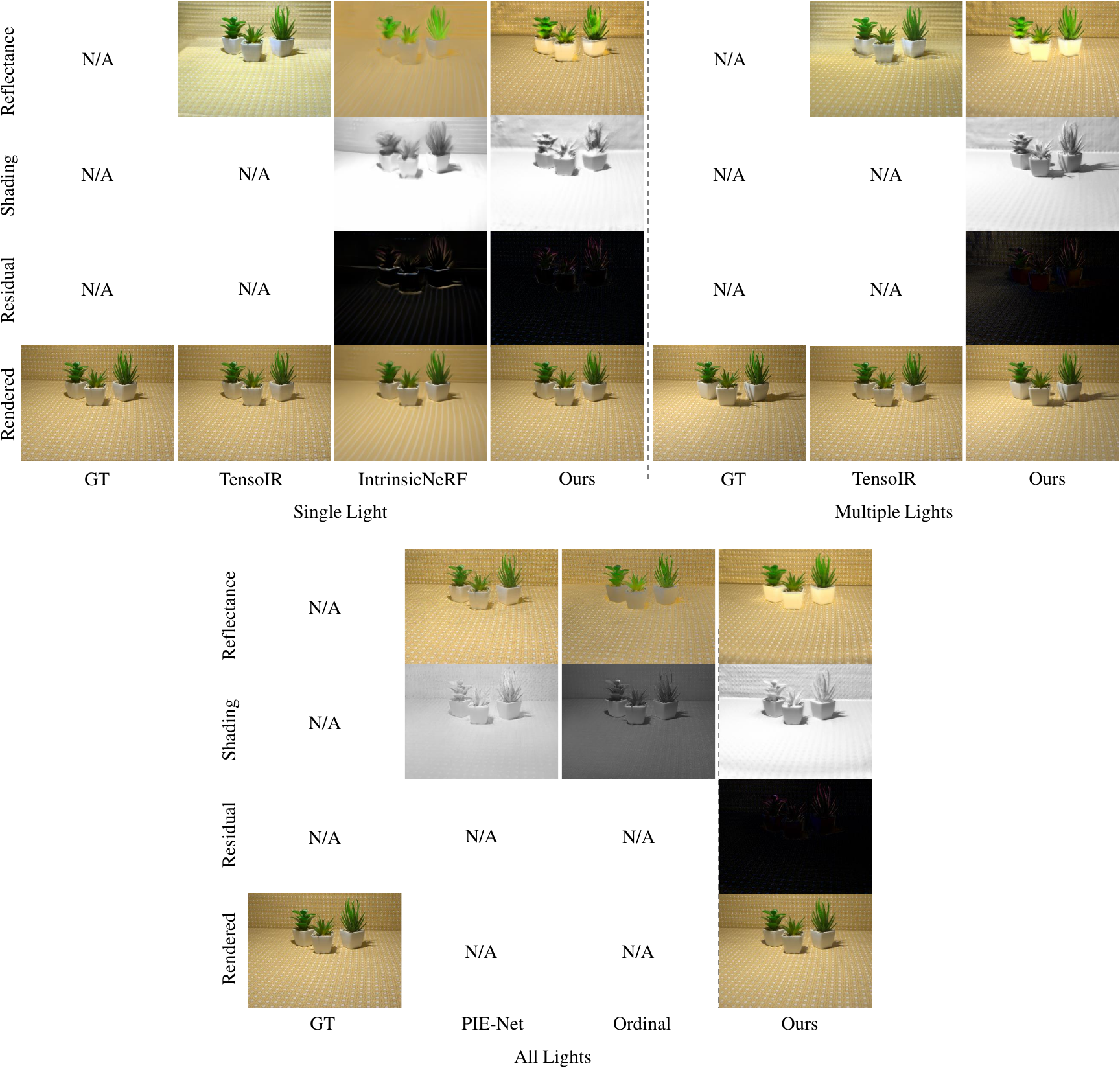}
\caption{Additional Qualitative Results on the ReNe dataset. (Garden).}
\label{suprene: Garden}
\end{figure*}

\begin{figure*}[!ht]
\centering
\includegraphics[width=1.0\linewidth]{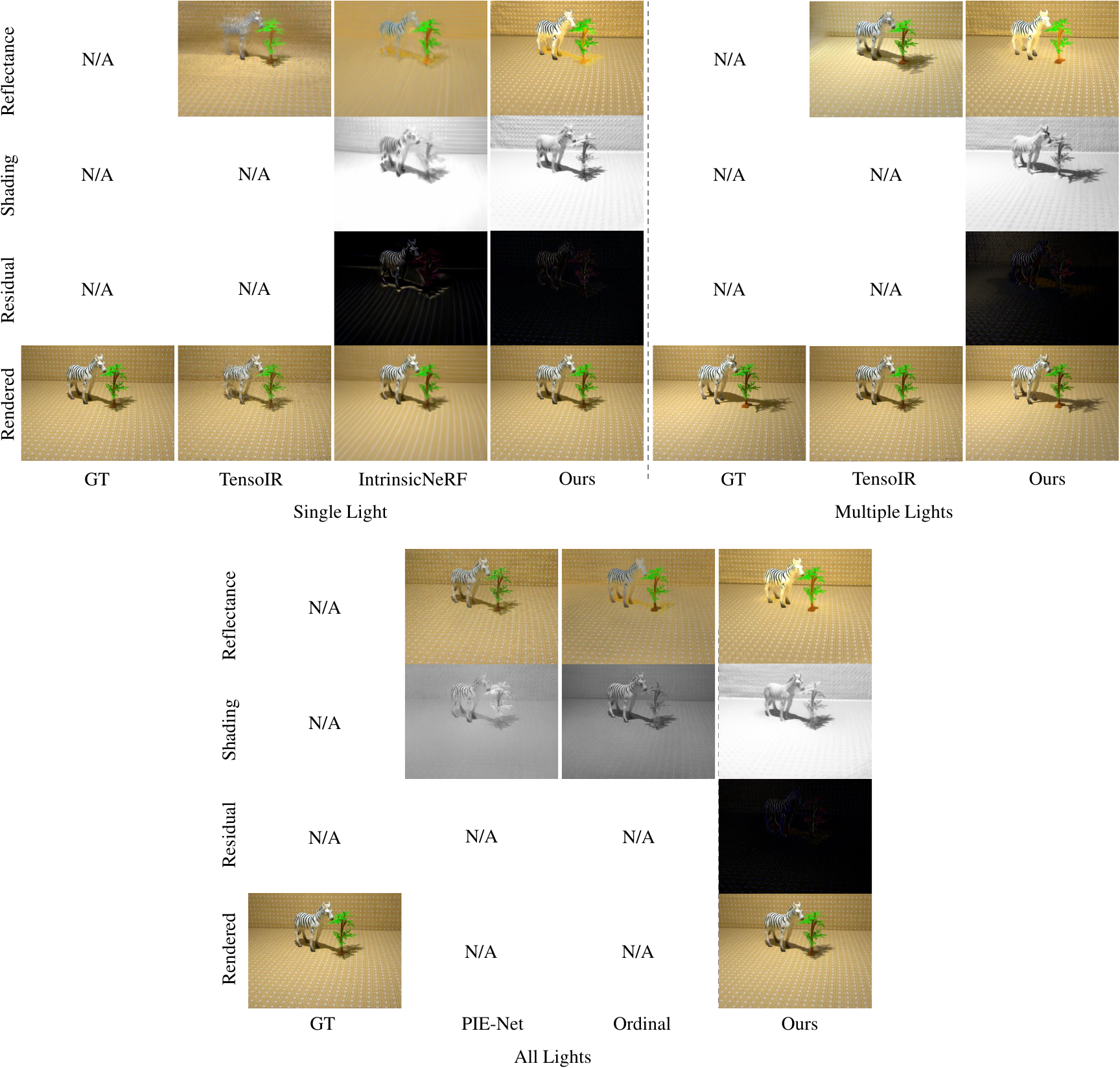}
\caption{Additional Qualitative Results on the ReNe dataset. (savannah).}
\label{suprene: savannah}
\end{figure*}

\cref{suprene: Apple} to \cref{suprene: savannah} present more detailed results on the ReNe dataset, where the four scenes are labeled as apple, cube, garden, and savannah in the original dataset. Similar to our observations on the Synthetic Dataset, our method outperforms previous methods across all settings, including the original ALL Lights, as well as the additional Single Light and Multiple Lights settings. Furthermore, the performance of our method improves as the number of light sources increases. In all scenes under the ALL Lights and Multiple Lights settings, our method consistently produces satisfactory intrinsic images.

The ReNe dataset poses significant challenges for 3D reconstruction and intrinsic decomposition due to the camera views and light views being concentrated within a limited area. As shown in Fig. \ref{suprene: Apple}, TensoIR fails to produce valid outputs in the Apple scene under both the Single Light and Multiple Lights settings. Despite these challenges, our method consistently delivers satisfactory results.

\clearpage
\clearpage

\end{document}